\documentclass{article}
\usepackage[T1]{fontenc}
\usepackage{iclr2027_conference,times}

\usepackage{amsmath,amsfonts,bm}

\def\eqref#1{equation~\ref{#1}}

\def\1{\bm{1}}

\DeclareMathAlphabet{\mathsfit}{\encodingdefault}{\sfdefault}{m}{sl}
\SetMathAlphabet{\mathsfit}{bold}{\encodingdefault}{\sfdefault}{bx}{n}

\usepackage{amsmath,amssymb}
\usepackage{booktabs}
\usepackage{array}
\usepackage{graphicx}
\usepackage{multirow}
\usepackage{microtype}
\usepackage{xcolor}
\usepackage{colortbl}
\usepackage{hyperref}
\usepackage{url}
\hypersetup{hidelinks}

\definecolor{cueblue}{HTML}{2878B5}
\definecolor{nuisanceorange}{HTML}{D97732}
\definecolor{fieldpurple}{HTML}{7B5BA7}
\definecolor{outputgreen}{HTML}{2A936B}
\definecolor{softgray}{HTML}{F2F4F7}
\definecolor{rankfirst}{HTML}{174A7E}
\definecolor{ranksecond}{HTML}{4078A6}
\definecolor{rankthird}{HTML}{536879}
\definecolor{rankfirstbg}{HTML}{7EADD2}
\definecolor{ranksecondbg}{HTML}{BCD4E8}
\definecolor{rankthirdbg}{HTML}{EDF3F9}

\title{Learning from Automatic 3D Gaussian Splatting Edits for Online Bayesian Scene Change Detection}

\author{Changrui Zhu {\normalfont(\href{mailto:changrui.zhu.19@ucl.ac.uk}{changrui.zhu.19@ucl.ac.uk})} \& Simon Julier {\normalfont(\href{mailto:s.julier@ucl.ac.uk}{s.julier@ucl.ac.uk})}\\
University College London\\
\url{https://changruizhu96.github.io/work/lego-scd/}}

\iclrfinalcopy
\begin{document}
\maketitle
\fancyhead{}

\begin{abstract}
Online scene change detection requires integrating incoming observations to determine which regions have changed relative to a reference scene.
We update change beliefs as new images arrive using generalised Bayesian evidence fusion.
However, the prior probability of change and evidence models are unknown beforehand, and numerical Bayesian inference can be costly.
We address these challenges by learning offline from automatic edits to a reference 3D Gaussian Splatting (3DGS) model.
These edits generate labelled training views without manual annotations.
Together with unchanged reference observations, they provide supervision for learning scene-specific evidence models and estimating the effective pixel prior and response statistics.
Online prediction combines current evidence with historical support, while closed-form local Bayes factors, parameterized by the learned response statistics, update an accumulated 3D Gaussian field.
Object-guided refinement produces the final change masks.
Our method achieves state-of-the-art accuracy on PASLCD and CCS3D, with $67.64\%$ and $75.82\%$ mIoU, respectively.
On PASLCD, the complete pipeline runs at $17.2$\,FPS over complete query sequences, the fastest among the evaluated online methods.
\end{abstract}

\section{Introduction}
\label{sec:introduction}

Scene change detection identifies physical changes in an environment from images captured at different times.
When captures follow different camera trajectories, viewpoint and visibility differences complicate direct image comparison \citep{lu2025gs3dcd,wu2026scenediff}.
A reference 3D Gaussian Splatting (3DGS) model \citep{kerbl2023gaussians} enables rendering the earlier scene at query viewpoints, providing a common view for comparison.
Recent methods, such as MV3DCD \citep{galappaththige2025mv3dcd} and O-SCD \citep{galappaththige2026oscd}, integrate multi-view change cues by optimizing per-Gaussian change parameters, using batch and online processing, respectively.

In O-SCD, processing each new image requires iterative optimization using current and replayed observations, adding computational overhead during online inference.
However, updating predictions as new evidence arrives does not inherently require repeated parameter optimization.
This motivates updating change beliefs as new images arrive. Generalised Bayesian inference \citep{bissiri2016general} provides a framework to combine a prior probability of change with incoming evidence.

Putting this idea into practice requires a prior and models that translate image differences into change evidence, which are unknown beforehand.
We estimate these quantities offline using automatic reference 3DGS edits and unchanged observations.
The edits turn the reference representation into a source of labelled training views, supporting scene-specific learning without manual change annotations.
This reduces dependence on large labelled datasets for training a single evidence model across diverse scene appearances and changes.
The learned evidence models also support efficient Bayesian evidence accumulation.
Evaluating Bayesian evidence can require numerical integration or sampling \citep{kass1995bayesfactors}, adding substantial cost when repeated for incoming frames.
We apply the learned evidence models to unchanged reference observations to estimate response statistics, which parameterize a local Bayesian model that admits closed-form Bayes factors.
We combine current evidence with historical support for prediction and use these Bayes factors to update the accumulated field for subsequent queries.
These updates require neither online backpropagation nor past-image replay.
Finally, object-guided refinement sharpens boundaries and completes missed object regions.

Our method achieves state-of-the-art accuracy on PASLCD and CCS3D, with $67.64\%$ and $75.82\%$ mIoU, respectively.
On PASLCD, it outperforms O-SCD Online by $18.40$ mIoU points while running at $17.2$\,FPS, compared with $14.4$\,FPS for O-SCD Online.
Without refinement, the pixel output achieves $60.36\%$ mIoU at $27.9$\,FPS.
Our complete method also exceeds GS-Diff's reported $64.40\%$ mIoU \citep{galappaththige2026gsdiff} without requiring a post-change reconstruction.
Given an existing reference reconstruction, automatic editing and model fitting take approximately 3.6 minutes per PASLCD instance on average.

\section{Related Work}
\label{sec:related}

\textbf{Scene change detection.}
Image-pair methods identify changes through correspondence and appearance comparison.
GeSCF uses SAM features and mask matching without task-specific training \citep{kim2025gescd}, while SceneDiff compares objects across before--after captures using semantic and geometric information \citep{wu2026scenediff}.
A reference 3D representation additionally enables comparisons at matched viewpoints and integration across observations.
3DGS-CD associates image-level detections across views \citep{lu2025gs3dcd}, and SCaR-3D combines change cues through multi-view voting and pruning \citep{zhou2026scar3d}.
MV3DCD optimizes Gaussian change channels from a collection of post-change images \citep{galappaththige2025mv3dcd}; O-SCD extends this approach to streaming queries, updating change parameters using current and replayed observations \citep{galappaththige2026oscd}.
CL-Splats uses detected changes to guide local updates of a Gaussian scene representation \citep{ackermann2025clsplats}.
GS-Diff instead compares geometry and colour between independently reconstructed scene states \citep{galappaththige2026gsdiff}.
Our method accumulates evidence on the reference Gaussians without online optimization or past-image replay.

\textbf{Learning from synthetic changes and 3D editing.}
CYWS constructs training pairs from object segmentation data \citep{sachdeva2023cyws}, while CYWS-3D generates paired observations from synthetic 3D scenes \citep{sachdeva2023cyws3d}.
Both use synthetic supervision to train change detectors that generalize to unseen scenes, requiring training data to cover diverse scene appearances and changes.
Advances in 3DGS segmentation \citep{ye2024gaussiangrouping,zhu2025unifiedlift}, editing \citep{chen2024gaussianeditor,lee2025editsplat}, and relighting \citep{liang2024gsir,gao2024relightable} enable controlled scene modifications.
RoboSplat edits Gaussian scenes and varies their appearance to train robot policies \citep{yang2025robosplat}, while \citet{zanjani2025gaussiandata} insert Gaussian objects into driving scenes to train 3D detectors.
Given a prebuilt reference 3DGS and its associated unchanged observations, we generate labelled changes through automatic edits and fit scene-specific evidence models and a prior offline.
This allows learning to focus on the appearance and response statistics of each reference scene, rather than requiring a single evidence model to cover many different environments.

\textbf{Bayesian evidence accumulation.}
Bayesian updates provide a way to combine successive observations with existing beliefs.
In scene mapping, POCD updates probabilistic object states using geometric and semantic information to track changes and maintain volumetric maps \citep{qian2022pocd}.
\citet{bissiri2016general} provide a generalised Bayesian framework that connects observations to belief updates through a loss function.
For online time-series changepoint detection, \citet{altamirano2023robust} construct generalised Bayesian updates with closed-form posterior parameters using diffusion score matching.
We apply generalised Bayesian prediction to scene change detection, using offline-learned response statistics to evaluate local Bayes factors in closed form for accumulation on reference Gaussians.


\section{Preliminaries}
\label{sec:preliminaries}

Given a reference 3D Gaussian Splatting (3DGS) model $\mathcal G$ and its registered reference images $\mathcal I_{\mathrm{ref}}$, we detect changes in incoming query images $I_t$ relative to the reference scene.
Comparing each query with the reference requires rendering the earlier scene from the query viewpoint.
Following O-SCD \citep{galappaththige2026oscd}, we first match XFeat keypoints \citep{potje2024xfeat} between the query and registered reference images to obtain 2D--3D correspondences, then estimate the query pose using PnP-RANSAC.
At the estimated pose, the reference 3DGS renders image $\widehat I_t$ through front-to-back alpha compositing \citep{kerbl2023gaussians}.
Each Gaussian $g\in\mathcal G$ contributes a view-dependent colour $\mathbf c_{tg}$ with projected opacity $\alpha_{tug}\in[0,1]$ at pixel $u$, giving $\widehat I_t(u)=\sum_g a_{tug}\mathbf c_{tg}$.
Here, $a_{tug}=\alpha_{tug}\prod_{h\prec_t g}(1-\alpha_{tuh})$ denotes its rendering contribution, with $h\prec_t g$ identifying Gaussians preceding it along the viewing ray.
These contributions also connect pixels to reference Gaussians for evidence aggregation in Section~\ref{sec:method}.
The reference reconstruction remains fixed during online inference.

Comparing $I_t$ with $\widehat I_t$ produces the initial change cues.
Following O-SCD, we compute a photometric difference from RGB $L_1$ and D-SSIM \citep{wang2004ssim}, and a semantic difference from the channel-averaged absolute difference between frozen SAM2 encoder features \citep{ravi2025sam2}.
Per-image normalization and upsampling of the semantic difference produce pixel-aligned cues $\widetilde D_t^{\mathrm{rgb}}$ and $\widetilde D_t^{\mathrm{sem}}$.
Their average gives the initial difference response $R_t^0=\tfrac12(\widetilde D_t^{\mathrm{rgb}}+\widetilde D_t^{\mathrm{sem}})\in[0,1]$, which serves as the input to our evidence learning and online inference.
Appendix~\ref{app:image-evidence} provides the detailed cue construction and normalization.

\section{Method}
\label{sec:method}

\subsection{Problem Formulation and Overview}
\label{sec:formulation}

Image differences mix physical changes with rendering artifacts and illumination shifts.
Our method uses generalised Bayesian inference to interpret these responses and accumulate change evidence.
For query image $I_t$, current-view evidence $C_t$ is constructed from the initial difference response $R_t^0$, while historical support $H_t$ summarizes evidence from earlier observations in the query view.
We formulate change prediction as a generalised Bayesian update \citep{bissiri2016general}: current and historical evidence modify the prior odds of change, determined by an effective pixel prior $\pi$.
The posterior change probability $Q_t^{\mathrm{pixel}}(u)\triangleq q_t(y=1\mid u)$, where $y=1$ denotes change, satisfies
\begin{equation}
\underbrace{\frac{Q_t^{\mathrm{pixel}}(u)}{1-Q_t^{\mathrm{pixel}}(u)}}_{\text{posterior odds}} =
\underbrace{\frac{\pi}{1-\pi}}_{\text{prior odds}}
\cdot
\underbrace{\frac{C_t(u)}{1-C_t(u)}}_{\text{current odds}}
\cdot
\underbrace{\frac{H_t(u)}{1-H_t(u)}}_{\text{history odds}}.
\label{eq:final-support}
\end{equation}
Neutral evidence, $C_t(u)=H_t(u)=0.5$, leaves $Q_t^{\mathrm{pixel}}(u)=\pi$.

\begin{figure}[t]
\centering
\includegraphics[width=\textwidth]{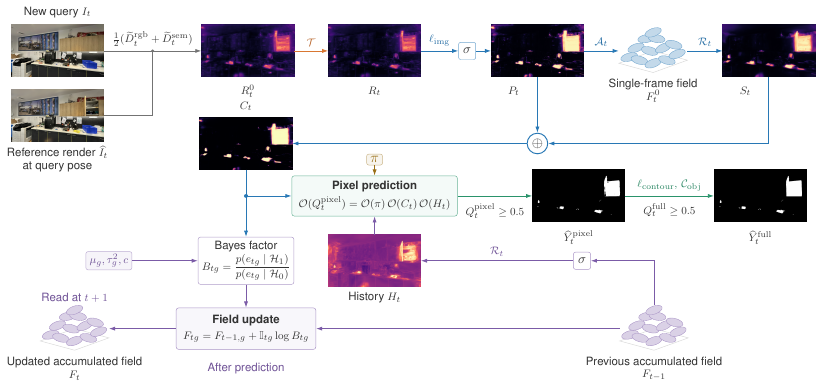}
\caption{\textbf{Online evidence fusion pipeline.}
Current evidence $C_t$ combines mapped image evidence $P_t$ and Gaussian-aggregated support $S_t$.
Fusion with prior $\pi$ and history $H_t$ from $F_{t-1}$ gives $Q_t^{\mathrm{pixel}}$; refinement produces $Q_t^{\mathrm{full}}$.
Thresholding at $0.5$ gives masks $\widehat Y_t^{\mathrm{pixel}}$ and $\widehat Y_t^{\mathrm{full}}$, respectively.
Local Bayes factors $B_{tg}$ update $F_{t-1}$ to $F_t$ for subsequent queries.
$\oplus$ adds log odds.}
\label{fig:online-fusion}
\end{figure}

However, the prior is unknown beforehand, and constructing $C_t$ and $H_t$ requires models that interpret image differences as change evidence and determine how this evidence contributes to history.
To obtain these components, we fit the effective pixel prior $\pi$ and four scene-specific models using supervision from automatic reference 3DGS edits and unchanged reference observations.
Of these four models, the response suppressor $\mathcal T$ and log-likelihood-ratio mapping $\ell_{\mathrm{img}}$ construct $C_t$, while the unchanged-response model $(\mu_g,\tau_g^2,c)$ supplies statistics for closed-form evidence accumulation underlying $H_t$.
The contour filter $\ell_{\mathrm{contour}}$ supports subsequent mask refinement.
All five components are fitted offline and remain fixed during online inference.
Section~\ref{sec:online-fusion} describes how they are used for prediction, history accumulation, and mask refinement (Figure~\ref{fig:online-fusion}), taking their fitted parameters as given.
Section~\ref{sec:learning} explains how automatic reference edits and unchanged observations are used to fit them (Figure~\ref{fig:pipeline}).

\subsection{Causal Online Change Detection}
\label{sec:online-fusion}

For each incoming query image $I_t$, we construct current-view evidence $C_t$, which supports both prediction and history accumulation. For prediction, $C_t$ combines with historical support $H_t$ from earlier frames; for accumulation, it provides the Gaussian responses used to compute local Bayes factors. The updated history is used only by subsequent queries.

\textbf{Current-view evidence.}
To suppress view-synthesis artifacts in initial response $R_t^0$ (Section~\ref{sec:preliminaries}), two predictors give change log odds $\ell_{\mathrm{base}}$ from response features and $\ell_{\mathrm{full}}$ from response and reference rendering features. The response suppressor $\mathcal T$ uses their difference to obtain corrected response $R_t$:
\begin{equation}
R_t = \sigma\!\left(\operatorname{logit} R_t^0 + \mathcal T(\ell_{\mathrm{base}},\ell_{\mathrm{full}})\right).
\label{eq:response-suppression}
\end{equation}
Here $\sigma(x)=(1+e^{-x})^{-1}$ is the sigmoid function and $\operatorname{logit}(p)=\log[p/(1-p)]$.
Although per-frame normalization unifies diverse cues, independent scaling leaves $R_t$ contrast-dependent. We therefore map $R_t$ to a shared evidence scale learned from reference data via the monotonic LLR mapping $\ell_{\mathrm{img}}$, giving probabilistic image evidence $P_t = \sigma(\ell_{\mathrm{img}}(R_t)) \in (0, 1)$, where $P_t = 0.5$ represents neutral evidence.
To suppress isolated pixel noise, we aggregate pixel evidence onto Gaussians and render it back to pixels using Gaussian blending weights $a_{tug}$. Specifically, aggregation $\mathcal A_t$ averages 2D pixel evidence $X_u$ within each Gaussian's footprint, while rendering $\mathcal R_t$ blends 3D Gaussian support $V_g$ back into pixels:
\begin{equation}
[\mathcal A_t(X)]_g = \frac{\sum_u a_{tug} X_u}{s_{tg}},
\qquad
[\mathcal R_t(V)]_u = \frac{\sum_g a_{tug} V_g}{\chi_{tu}},
\label{eq:render-operators}
\end{equation}
where $s_{tg} = \sum_u a_{tug}$ and $\chi_{tu} = \sum_g a_{tug}$ normalize the weights, preserving neutral evidence ($0.5$).
We apply these operators to obtain the single-frame response field $F_t^0 = \mathcal A_t(P_t)$ with entries in $[0,1]$, and Gaussian-aggregated support $S_t = \mathcal R_t(F_t^0)$. Combining $P_t$ with $S_t$ retains pixel detail while incorporating evidence shared through Gaussian footprints, giving current-view evidence $C_t$:
\begin{equation}
C_t = \sigma\!\left(\operatorname{logit} P_t + \operatorname{logit} S_t\right).
\label{eq:current-support}
\end{equation}

\textbf{3D evidence accumulation.}
\label{sec:neighbourhood}
To determine how current-view evidence $C_t$ should contribute to history, we evaluate Gaussian responses relative to their expected behaviour under unchanged conditions. The unchanged-response model provides each Gaussian's expected response $\mu_g$, the uncertainty in the estimated unchanged-response mean $\tau_g^2$, and the spatial co-response coefficient $c$. Each Gaussian's aggregated response $r_{tg} = [\mathcal A_t(C_t)]_g$ is standardized as $z_{tg} = (r_{tg}-\mu_g)/\sqrt{\tau_g^2+v_{tg}}$, where $v_{tg} = [\mathcal A_t(C_t^2)]_g-r_{tg}^2$ measures response variance within its footprint. The observation indicator $\mathbb I_{tg}=\mathbb 1\{s_{tg}\geq1\}$ selects Gaussians with sufficient rendering contribution.

A high response from one Gaussian may reflect local noise. Since object changes can affect several neighbouring Gaussians, we combine nearby responses using the neighbourhood mean $\bar z_{tg} = (\sum_j\eta_{gj}\mathbb I_{tj}z_{tj})/(\sum_j\eta_{gj}\mathbb I_{tj})$, where $j$ indexes neighbours of Gaussian $g$ on the reference 3D graph. Edge weights $\eta_{gj}$ reflect spatial and colour similarity (Appendix, Eq.~\ref{eq:app-affinity}), and $\mathbb I_{tj}$ applies the same observation rule to neighbour $j$.
Neighbours can also respond together without change. Under the unchanged hypothesis $\mathcal H_0$, we model the central response $z_{tg}$ and neighbourhood mean $\bar z_{tg}$ as jointly Gaussian with zero means, unit central variance, and covariance approximated by $c$. This gives the conditional expectation $\mathbb E[\bar z_{tg}\mid z_{tg},\mathcal H_0]\approx cz_{tg}$ (Appendix, Eq.~\ref{eq:app-conditional-mean}). Subtracting it yields the residual $e_{tg}=\bar z_{tg}-cz_{tg}$, which measures neighbour response beyond that expected from the centre.

To quantify this residual as change evidence, we model $e_{tg}\mid\mathcal H_0\sim\mathcal N(0,\nu_{tg})$. The residual variance $\nu_{tg}=\max(V_{tg}-c^2,\epsilon_\nu)$ uses the estimated variance $V_{tg}$ of the neighbourhood mean (Appendix, Eq.~\ref{eq:app-neighbour-variance}); $\epsilon_\nu=10^{-4}$ prevents degenerate variance. Under the change hypothesis $\mathcal H_1$, a shared response increase $\delta\geq0$ at the centre and its neighbours gives $e_{tg}\mid\delta,\mathcal H_1\sim\mathcal N((1-c)\delta,\nu_{tg})$, with $\delta\sim\operatorname{HalfNormal}(1)$ and density $p_+(\delta)=2\mathcal N(\delta;0,1)$ for $\delta\geq0$.
We integrate over $\delta$ to evaluate evidence for change. The Gaussian likelihood and half-normal prior give an exponent quadratic in $\delta$, so completing the square evaluates this integral analytically. With $h_{tg}=(1-c)e_{tg}/\nu_{tg}$ and $J_{tg}=(1-c)^2/\nu_{tg}+1$, the local Bayes factor $B_{tg}$ is
\begin{equation}
B_{tg}\!=\!\frac{p(e_{tg}\mid\mathcal H_1)}{p(e_{tg}\mid\mathcal H_0)}
\!=\!\frac{\displaystyle\int_0^\infty\mathcal N\!\left(e_{tg};(1-c)\delta,\nu_{tg}\right)p_+(\delta)\,d\delta}{\mathcal N(e_{tg};0,\nu_{tg})}
\!=\!\frac{2}{\sqrt{J_{tg}}}
\exp\!\left(\frac{h_{tg}^2}{2J_{tg}}\right)
\Phi\!\left(\frac{h_{tg}}{\sqrt{J_{tg}}}\right),
\label{eq:conditional-evidence}
\end{equation}
where $\Phi$ is the standard normal cumulative distribution function. This expression requires neither numerical integration nor iterative optimization; its derivation and log form are given in Appendix~\ref{app:conditional}, Eqs.~\ref{eq:app-bf-integral}--\ref{eq:app-bf-closed}.
Values $B_{tg}>1$ support change and $B_{tg}<1$ oppose it, with neutral evidence $B_{tg}=1$ when no neighbours are observed. Adding these log Bayes factors to the previous accumulated field $F_{t-1}$ gives the updated log-evidence field $F_t$ for subsequent queries, with real-valued entries $F_{tg}$ at Gaussians $g$:
\begin{equation}
F_{tg} = F_{t-1,g} + \mathbb I_{tg} \log B_{tg}.
\label{eq:field-update}
\end{equation}

\textbf{Historical fusion and causal prediction.}
\label{sec:history}
To predict pixel-level changes on query $I_t$, historical support $H_t$ is rendered from the previous accumulated field $F_{t-1}$ (with $F_0 = 0$):
\begin{equation}
H_t = \mathcal R_t\!\left(\sigma(F_{t-1})\right),
\label{eq:history-support}
\end{equation}
where $\sigma$ maps accumulated log evidence into $(0,1)$. Rendering accounts for view-dependent visibility and occlusion; unobserved Gaussians ($F=0$) give neutral support $0.5$ on covered pixels. Using Equation~\ref{eq:final-support}, we combine $C_t$ from Equation~\ref{eq:current-support}, $H_t$ above, and the effective pixel prior $\pi$ fitted offline. Converting the resulting odds to probabilities gives the pixel scores $Q_t^{\mathrm{pixel}}$. The binary change mask is $\widehat Y_t^{\mathrm{pixel}}(u)=\mathbb{1}\{Q_t^{\mathrm{pixel}}(u)\geq0.5\}$.

\textbf{Object-guided mask refinement.}
Gaussian overlap and alpha-blending can blur rendered change boundaries. We use connected components of $\widehat Y_t^{\mathrm{pixel}}$ to prompt the SAM2 decoder on both query and reference renderings. 3DGS-CD~\citep{lu2025gs3dcd} fuses prompted segmentations into 3D object masks; our contour refinement retains the original scores of unprompted pixels. The contour model predicts log odds $\ell_{\mathrm{contour}}(u)$ from paired mask features. Their non-positive part penalizes pixel change scores, giving the contour-filtered score $Q_t^{\mathrm{contour}}(u)$:
\begin{equation}
Q_t^{\mathrm{contour}}(u) = \sigma\!\left(\operatorname{logit} Q_t^{\mathrm{pixel}}(u) + \min(\ell_{\mathrm{contour}}(u),\, 0)\right).
\label{eq:contour-refinement}
\end{equation}
To recover missed object parts, surviving components prompt SAM2 on the query for candidate masks $M$. Their mean log-odds evidence is $\overline L_M = \operatorname{mean}_{u \in M} \operatorname{logit} Q_t^{\mathrm{pixel}}(u)$. We retain masks with $\overline L_M \ge 0$ and raise covered pixels to at least their mask score, giving the refined change score $Q_t^{\mathrm{full}}(u)$:
\begin{equation}
Q_t^{\mathrm{full}}(u) = \max\!\left(Q_t^{\mathrm{contour}}(u),\; \max_{M \ni u} \sigma(\overline L_M)\right),
\label{eq:refined-output}
\end{equation}
where the inner maximum is over retained masks covering $u$ and is set to $0$ if none cover it. Thresholding at $0.5$ gives the refined mask $\widehat Y_t^{\mathrm{full}}$ (Appendix~\ref{app:completion}).

\subsection{Learning from Automatic 3DGS Edits}
\label{sec:learning}

\begin{figure}[t]
\centering
\includegraphics[width=\textwidth]{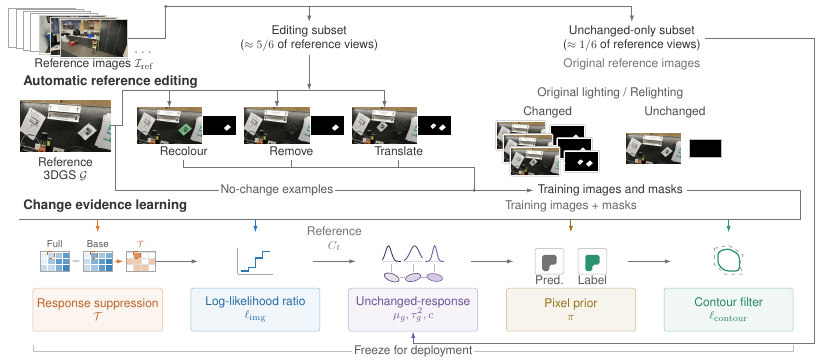}
\caption{\textbf{Automatic reference editing and change evidence learning.}
\textbf{Top}: Reference 3DGS edits (recolouring, removal, translation) and relighting synthesize labelled change pairs without manual annotation.
\textbf{Bottom}: Five scene-specific components are fitted in the order required by their inputs (grey arrows): response suppressor $\mathcal T$, LLR mapping $\ell_{\mathrm{img}}$, unchanged-response model $(\mu_g, \tau_g^2, c)$, pixel prior $\pi$, and contour filter $\ell_{\mathrm{contour}}$.}
\label{fig:pipeline}
\end{figure}

Figure~\ref{fig:pipeline} illustrates our framework for learning from automatic 3DGS edits: we first synthesize object-level edits on the reference 3DGS to generate labelled training views, and then fit the scene-specific components in the order required by their inputs.

\textbf{Automatic reference editing.}
\label{sec:matched-data}
We split reference cameras approximately $5{:}1$ into an \textit{editing subset} for learning and an \textit{unchanged-only subset} for estimating unchanged-response statistics. A regular grid of automatic point prompts drives the SAM2 mask decoder \citep{ravi2025sam2} on reference views, and masks are matched across views through their contributing Gaussians. Aggregating matched masks $X$ via $[\mathcal A_t(X)]_g$ (Eq.~\ref{eq:render-operators}) yields object membership $m_g \in [0, 1]$, measuring the fraction of each Gaussian's rendering footprint inside the object. We select objects based on their consistency across held-out views. Recolouring blends colours towards the palette colour furthest from the object's median colour with mixing coefficient $0.45m_g$; removal scales opacity by $1 - m_g$; translation applies a shared 3D displacement to Gaussians with $m_g \ge 0.5$. Edits modify duplicate models, preserving the original reference.
For reference image $I_t^{\mathrm{ref}}\in\mathcal I_{\mathrm{ref}}$, original and edited models render $\widehat I_t$ and $\widehat I_t^{\mathrm{edit}}$ at its registered pose; here $t$ indexes reference views. Synthetic query $I_t^{\mathrm{syn}} = I_t^{\mathrm{ref}} + (\widehat I_t^{\mathrm{edit}} - \widehat I_t)$ transfers changes to real reference images while preserving differences between reference images and their renderings. Regions covered by edit masks are labelled changed ($1$); valid regions outside are labelled unchanged ($0$), with ambiguous boundaries ignored. We introduce lighting variations using affine colour transforms, optionally augmented with normal-dependent directional shading, fitted to differences between unchanged reference images and their renderings. These transformations preserve the change labels (details and additional examples in Appendix~\ref{app:reference-edits}, Figure~\ref{fig:reference-edit-examples}).

\textbf{Response suppression.}
We fit the response suppressor $\mathcal T$ using two pixel-wise logistic classifiers on the editing subset. \textit{Base} uses initial response $R_t^0$ and its local texture statistics (mean, standard deviation, gradient). \textit{Full} adds pixel-aligned reference rendering features at the same pose: coverage, depth, Gaussian contribution statistics, and their gradients and interactions. The fitted classifiers predict change log odds $\ell_{\mathrm{base}}$ and $\ell_{\mathrm{full}}$. We suppress responses when adding rendering features lowers the predicted change odds, using
\begin{equation}
\mathcal T(\ell_{\mathrm{base}},\ell_{\mathrm{full}}) = \min(\ell_{\mathrm{full}} - \ell_{\mathrm{base}}, 0).
\label{eq:suppression-rule}
\end{equation}
Fitting details and feature groups are provided in Appendix~\ref{app:image-evidence} (Eq.~\ref{eq:app-nuisance}).

\textbf{Log-likelihood-ratio mapping.}
\label{sec:current}
Under Bayesian hypothesis testing, the corrected response $R_t$ is mapped to evidence via the log-likelihood ratio (LLR) between change ($\mathcal H_1$) and no-change ($\mathcal H_0$). To estimate this mapping, we pool corrected responses across the editing subset and use isotonic regression \citep{zadrozny2002calibration} to partition responses into monotonic blocks $b$. The unchanged distribution $p(b \mid \mathcal H_0)$ is an equal mixture of distributions estimated from regions outside edits ($\mathcal H_0^{\mathrm{edit}}$) and unedited reference views ($\mathcal H_0^{\mathrm{ref}}$). This gives the block LLR:
\begin{equation}
\ell_b = \log\frac{\widehat p(b\mid \mathcal H_1)}{\widehat p(b\mid \mathcal H_0)} = \log\frac{\widehat p(b\mid \mathcal H_1)}{\tfrac12\widehat p(b\mid \mathcal H_0^{\mathrm{edit}}) + \tfrac12\widehat p(b\mid \mathcal H_0^{\mathrm{ref}})}.
\label{eq:likelihood-ratio}
\end{equation}
Linearly interpolating $\ell_b$ across block-mean responses yields the continuous monotonic mapping $\ell_{\mathrm{img}}$ (Appendix~\ref{app:image-evidence}).

\textbf{Unchanged-response modelling.}
To fit the unchanged-response model $(\mu_g, \tau_g^2, c)$, we estimate its response statistics over reference views $t$ in the unchanged-only subset. We place a Beta prior on each Gaussian's latent mean response and treat bounded observations $r_{tg} \in [0,1]$ as fractional counts. Taking the prior concentration to zero from above yields the mean $\mu_g$ and uncertainty in the estimated unchanged-response mean $\tau_g^2$. The reference graph is built from eight-nearest-neighbour candidates, then symmetrized and weighted by spatial and colour similarity $\eta_{gj}$. Co-observed neighbour pairs estimate the spatial co-response coefficient $c$:
\begin{equation}
\mu_g = \frac{\sum_t \mathbb I_{tg}r_{tg}}{\sum_t \mathbb I_{tg}},\qquad
\tau_g^2 = \frac{\mu_g(1-\mu_g)}{1+\sum_t \mathbb I_{tg}},\qquad
c = \frac{\sum_{t,(g,j)}\omega^{\mathrm{fit}}_{tgj}\widetilde z_{tg}\widetilde z_{tj}}{\sum_{t,(g,j)}\omega^{\mathrm{fit}}_{tgj}},
\label{eq:unchanged-moments}
\end{equation}
where $\widetilde z_{tg}$ and $\widetilde z_{tj}$ denote responses of Gaussian $g$ and neighbour $j$ standardized across unchanged reference views, $\omega^{\mathrm{fit}}_{tgj} = \eta_{gj} \min(\mathbb I_{tg}, \mathbb I_{tj})$ weights co-observed neighbour pairs, and unobserved Gaussians ($\sum_t \mathbb I_{tg}=0$) revert to scene-level priors (details in Appendix~\ref{app:normal}, Eqs.~\ref{eq:app-beta}--\ref{eq:app-beta-moments}).

\textbf{Pixel prior fitting.}
\label{sec:decision}
To fit the pixel prior $\pi$ (Eq.~\ref{eq:final-support}) robustly against synthetic edit artifacts (e.g., exposed backgrounds or spillover), we introduce Cauchy reliability weights $\rho_i \in (0, 1]$ to downweight views with abnormal responses in labelled unchanged regions. For each training image $i$, let $Y_i$ denote its binary change-label mask and $\widehat Y_i^{\mathrm{pixel}}(\pi)$ the pixel prediction obtained with prior $\pi$. We determine the optimal prior $\pi^\star$ by maximizing their reliability-weighted foreground IoU over the editing subset:
\begin{equation}
\pi^\star \in \arg\max_{\pi \in (0, 1)} \frac{\sum_i \rho_i \operatorname{IoU}\!\left(\widehat Y_i^{\mathrm{pixel}}(\pi),\, Y_i\right)}{\sum_i \rho_i}.
\label{eq:decision-objective}
\end{equation}
This objective fits $\pi$ as an effective prior that sets how much evidence is needed to predict change.
We solve this one-dimensional search via branch-and-bound (Appendix~\ref{app:decision}, Eqs.~\ref{eq:app-search-bound}--\ref{eq:app-search-stop}).

\textbf{Contour filter fitting.}
To learn the boundary penalty in Eq.~\ref{eq:contour-refinement}, we first fit a linear classifier with score $s_{\mathrm{contour}}(u) = \mathbf w^\top \mathbf f_u + b$ over standardized features $\mathbf f_u$ that are unchanged when the query and reference images are swapped. These view-symmetric features combine SAM2 mask logits, predicted IoUs, and their products using $(\max, \min)$ across the two views. Across training image pairs $i$ and candidate pixels $u$, the feature vectors are $\mathbf f_{iu}$ and edit-mask labels are $y_{iu}=Y_i(u)\in\{0,1\}$. We fit $\mathbf w$ and $b$ by minimizing regularized binary cross-entropy:
\begin{equation}
\min_{\mathbf w, b} \sum_{i, u} w_{iu} \mathcal L_{\mathrm{CE}}\!\left(\sigma(\mathbf w^\top\mathbf f_{iu}+b),\, y_{iu}\right) + \frac{\lambda}{2} \|\mathbf w\|_2^2,
\label{eq:contour-fitting}
\end{equation}
where $\lambda$ controls $L_2$ regularization. Excluding ignored pixels, we construct $w_{iu}$ by balancing classes within each image, multiplying by its reliability $\rho_i$, and normalizing each class to total weight $0.5$ within the fitting split. Platt scaling \citep{platt2000probabilities} fits a slope $a_{\mathrm{cal}}$ and intercept $b_{\mathrm{cal}}$ on held-out objects using the same weighting procedure, giving contour log odds $\ell_{\mathrm{contour}}(u)=a_{\mathrm{cal}}s_{\mathrm{contour}}(u)+b_{\mathrm{cal}}$ (Appendix~\ref{app:completion}).

\section{Experiments}
\label{sec:experiments}

\noindent\textbf{Protocol.}
Real-world PASLCD \citep{galappaththige2025mv3dcd} is our primary benchmark (20 scene-instances, conventional $4\times$ downsampling), complemented by evaluation on four synthetic CCS3D scenes \citep{zhou2026scar3d}. Following O-SCD's mask definition, we evaluate our method and O-SCD on both datasets using masks covering pre- and post-state changes, including vacated and newly occupied regions. Tables~\ref{tab:overall}--\ref{tab:ccs3d} compare our method with official-code O-SCD results and other published online/offline baselines. We report foreground mIoU and F1 for \textbf{Ours (pixel)} ($\widehat Y_t^{\mathrm{pixel}}$) and \textbf{Ours (full)} ($\widehat Y_t^{\mathrm{full}}$) at threshold $0.5$ (Figure~\ref{fig:online-fusion}). Models are fitted offline from reference data without test-time tuning. Our comparisons average two complete runs per dataset (Appendix~\ref{app:updated-evaluation}).

\begin{figure}[t]
\centering
\includegraphics[width=\linewidth]{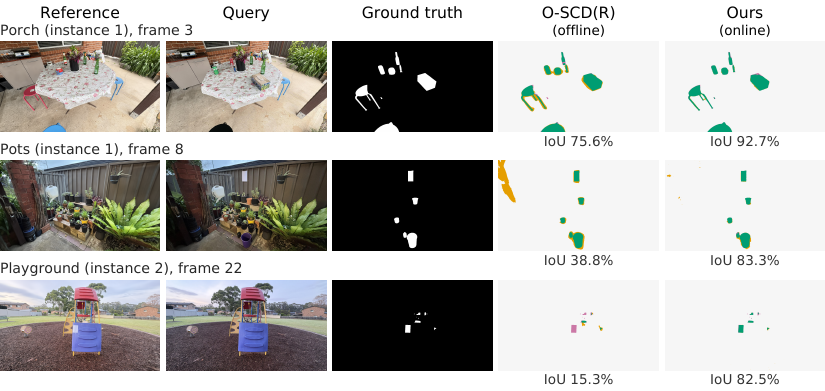}
\caption{\textbf{Visual results on representative scenes.}
Columns show reference renderings, queries, ground truth, O-SCD(R) (offline),
and our complete method (online).
\textcolor[RGB]{0,158,115}{\rule{0.8em}{0.8em}}~correct detections;
\textcolor[RGB]{230,159,0}{\rule{0.8em}{0.8em}}~false positives;
\textcolor[RGB]{204,121,167}{\rule{0.8em}{0.8em}}~missed changes.
Numbers give per-frame foreground IoU from one complete sequence run.
Additional visual results, including remaining errors, are shown in Appendix~\ref{app:visual-results}.}
\label{fig:qualitative}
\end{figure}
\begin{table}[t]
\centering
\begin{minipage}[t]{0.485\textwidth}
\vspace{0pt}
\centering
\caption{PASLCD (20 scene-instances, \%). Our method and O-SCD are evaluated locally; other results are taken from GS-Diff.\strut}
\label{tab:overall}
\small
\setlength{\tabcolsep}{1.5pt}
\begin{tabular*}{\linewidth}{@{\extracolsep{\fill}}lllrr@{}}
\toprule
Method & Venue & Mode & mIoU $\uparrow$ & F1 $\uparrow$\\
\midrule
\multicolumn{5}{l}{\textit{Paper-reported}}\\
SCaR-3D & 3DV 2026 & offline & 19.10 & 28.90\\
3DGS-CD & RA-L 2025 & offline & 20.90 & 33.90\\
MV3DCD & CVPR 2025 & offline & 47.80 & 62.80\\
GS-Diff & arXiv 2026 & offline & 64.40 & 75.80\\
\midrule
\multicolumn{5}{l}{\textit{Our evaluation}}\\
O-SCD(O) & CVPR 2026 & online & 49.24 & 64.60\\
O-SCD(R) & CVPR 2026 & offline & 55.99 & 70.27\\
Ours (pixel) & --- & online & 60.36 & 73.80\\
Ours (full) & --- & online & \textbf{67.64} & \textbf{78.87}\\
\bottomrule
\end{tabular*}
\end{minipage}%
\hfill%
\begin{minipage}[t]{0.485\textwidth}
\vspace{0pt}
\centering
\caption{CCS3D (four scenes, \%). Our method and O-SCD are evaluated locally; other results are taken from SCaR-3D.\strut}
\label{tab:ccs3d}
\small
\setlength{\tabcolsep}{1.5pt}
\begin{tabular*}{\linewidth}{@{\extracolsep{\fill}}lllrr@{}}
\toprule
Method & Venue & Mode & mIoU $\uparrow$ & F1 $\uparrow$\\
\midrule
\multicolumn{5}{l}{\textit{Paper-reported}}\\
MV3DCD & CVPR 2025 & offline & 30.40 & 44.10\\
CL-Splats & ICCV 2025 & offline & 39.90 & 53.80\\
3DGS-CD & RA-L 2025 & offline & 41.60 & 51.20\\
SCaR-3D & 3DV 2026 & offline & 65.00 & 72.40\\
\midrule
\multicolumn{5}{l}{\textit{Our evaluation}}\\
O-SCD(O) & CVPR 2026 & online & 49.94 & 64.61\\
O-SCD(R) & CVPR 2026 & offline & 43.59 & 57.82\\
Ours (pixel) & --- & online & 66.83 & 79.47\\
Ours (full) & --- & online & \textbf{75.82} & \textbf{85.90}\\
\bottomrule
\end{tabular*}
\end{minipage}
\end{table}

\noindent\textbf{Complete evaluation.}
Our method achieves state-of-the-art accuracy on PASLCD with $67.64\%$ mIoU and $78.87\%$ F1 (Table~\ref{tab:overall}), exceeding O-SCD(O) by $18.40$ mIoU points and offline GS-Diff \citep{galappaththige2026gsdiff} by $3.24$ points without a post-change reconstruction. Ours (pixel) also exceeds both O-SCD variants.
On CCS3D, our method exceeds O-SCD(O) by $25.88$ mIoU points under the same union-mask evaluation (Table~\ref{tab:ccs3d}).
Figure~\ref{fig:qualitative} shows more complete object masks in \emph{Porch}, fewer boundary false positives in \emph{Pots}, and fewer missed changes in \emph{Playground} than O-SCD(R). Appendix~\ref{app:additional-results} provides per-instance results.

\noindent\textbf{Cross-instance reference reuse.}
We reuse each reference reconstruction and its fitted parameters to process the other instance of the same PASLCD scene, without refitting. Overall pixel and full-output mIoU remain close to the original pairing: $59.93\%$ versus $60.36\%$, and $67.15\%$ versus $67.64\%$, respectively. These results support reference reuse between closely related instances of the same scene. Appendix~\ref{app:cross-instance} provides the protocol and per-pair results (Table~\ref{tab:cross-instance}).

\begin{table}[t]
\centering
\caption{Independent component ablations on 20 PASLCD instances (\%). Each component is removed separately from the corresponding complete pipeline with the decision threshold fixed at $0.5$. Removing the pixel prior is equivalent to using a neutral prior, $\pi=0.5$.}
\label{tab:component-ablation}
\small
\setlength{\tabcolsep}{3.5pt}
\begin{tabular*}{\linewidth}{@{\extracolsep{\fill}}lrr@{\hskip 16pt}lrr@{}}
\toprule
\multicolumn{3}{c}{\textit{Pixel pipeline}} & \multicolumn{3}{c}{\textit{Output refinement}}\\
\cmidrule(lr){1-3}\cmidrule(lr){4-6}
Configuration & mIoU $\uparrow$ & F1 $\uparrow$ & Configuration & mIoU $\uparrow$ & F1 $\uparrow$\\
\midrule
\textbf{Full pixel pipeline} & \textbf{60.36} & \textbf{73.80} & \textbf{Full method} & \textbf{67.64} & \textbf{78.87}\\
w/o response suppression & 58.62 & 72.17 & w/o contour filtering & 63.15 & 75.78\\
w/o likelihood-ratio mapping & 59.23 & 73.12 & w/o object completion & 62.56 & 75.32\\
w/o Gaussian-aggregated support & 58.47 & 72.27 & w/o contour image weights & 67.38 & 78.67\\
w/o historical support & 57.81 & 71.67 & & & \\
w/o pixel prior & 43.90 & 58.26 & & & \\
\bottomrule
\end{tabular*}
\end{table}

\noindent\textbf{Independent component ablations.}
Removing any evaluated component reduces mean mIoU (Table~\ref{tab:component-ablation}), with the pixel prior $\pi$ having the largest effect in the pixel pipeline: its removal causes a $16.47$-point drop. This highlights the importance of using reference data to determine how much evidence is needed to predict change. Appendix~\ref{app:prior-area} reports the fitted priors and performance by change area. Gaussian-aggregated support contributes $1.89$ points, supporting the use of spatially shared evidence. Response suppression and likelihood-ratio mapping provide additional gains.
Historical support connects each prediction to a shared 3D field that accumulates change evidence from earlier views. Removing it reduces mean pixel mIoU by $2.56$ points. To evaluate how this shared evidence should be accumulated, we replace our Bayes-factor updates with a temporal average of each Gaussian's observed responses, reducing pixel mIoU by $4.42$ points (Appendix, Table~\ref{tab:history-rules}).
Contour filtering suppresses excess responses, while object completion recovers missed object regions. Removing them reduces mean mIoU by $4.49$ and $5.08$ points, respectively, while contour image weights provide a smaller gain of $0.26$ points.

\noindent\textbf{Relighting strength.}
Removing the relighting during data generation reduces full-output mIoU by $0.66$ points, while doubling its strength gives similar aggregate results (Appendix, Table~\ref{tab:relighting-strength}).

\noindent\textbf{Runtime.}
We measure runtime on an NVIDIA RTX 5090 over all 25 frames of all 20 PASLCD instances after initialization (Table~\ref{tab:runtime}). Including frontend processing, our complete method reaches $17.2$\,FPS, compared with $14.4$\,FPS for O-SCD(O), while improving mIoU by $18.40$ points. Without refinement, our pixel pipeline reaches $27.9$\,FPS at $60.36\%$ mIoU, still exceeding the offline refined O-SCD(R) result of $55.99\%$. The learned models enable efficient online evidence accumulation through closed-form local Bayes factors, avoiding online backpropagation and past-image replay.
Given an existing reference reconstruction, offline preparation averages $215.82$\,s ($\sim$3.6\,minutes) per instance, including automatic editing and model fitting. Appendices~\ref{app:updated-evaluation} and~\ref{app:offline-timing} provide the timing protocol and offline cost breakdown.

\begin{table}[t]
\centering
\caption{Runtime on PASLCD (RTX 5090; 20 instances). Post-frontend covers detection, state updates, and refinement. Timing definitions and protocols are given in Appendix~\ref{app:updated-evaluation}.}
\label{tab:runtime}
\small
\setlength{\tabcolsep}{8pt}
\begin{tabular}{llrrr}
\toprule
Method & Mode & Post-frontend (ms) $\downarrow$ & End-to-end (ms) $\downarrow$ & FPS $\uparrow$\\
\midrule
O-SCD(R)$^{\dagger}$ & offline & 169.41 & 201.34 & 4.97\\
O-SCD(O) & online & 37.65 & 69.33 & 14.42\\
\midrule
Ours (pixel) & online & \textbf{4.79} & \textbf{35.81} & \textbf{27.93}\\
Ours (full) & online & 25.64 & 58.23 & 17.17\\
\bottomrule
\end{tabular}
\par\smallskip
\begin{minipage}{\linewidth}
\footnotesize $^{\dagger}$O-SCD(R) requires the full sequence; total runtime, including refinement, is reported per frame.
\end{minipage}
\end{table}

\section{Conclusion and Limitations}
\label{sec:limitations}

Our method learns scene-specific evidence models and statistics from automatic
reference 3DGS edits, using the reference scene as a source of supervision for
online change detection. These learned quantities enable closed-form local
Bayes factors and generalised Bayesian history accumulation without online
backpropagation or past-image replay. It outperforms O-SCD on PASLCD and CCS3D,
achieving $67.64\%$ mIoU at $17.2$\,FPS on PASLCD.
However, our direct object edits and simple relighting prioritize short offline
preparation and may not cover real changes or unusual lighting conditions.
Accumulating evidence on the unchanged reference geometry limits how accurately
newly introduced objects and occlusion boundaries can be represented in the field.
The response uncertainty and spatial covariance models are working approximations.
Although local Bayes factors are evaluated analytically under the residual model,
the fused scores combine dependent evidence and are not guaranteed to be
calibrated probabilities of real-world changes.
Future work will explore more advanced 3DGS segmentation, editing, and relighting
to improve object masks and broaden the generated changes, while assessing whether
the resulting detection gains justify the additional offline preparation time.

\section*{AI Use Statement}
In this work, generative AI tools were used to assist with initial methodology discussions and exploratory brainstorming, algorithm and code implementation, preliminary manuscript drafting, and textual refinement for grammar, phrasing, and readability. Generative AI tools were not used for formulating theoretical claims, deriving mathematical formulations, collecting benchmark observations, or interpreting empirical results.
All AI-assisted contributions were rigorously reviewed, tested, and vetted by the authors: all algorithmic implementations and evaluation pipelines were independently executed and verified, all mathematical derivations were checked manually, and all experimental results and claims were thoroughly audited. The authors assume full responsibility for the entirety of this paper and its claims.

\section*{Ethics Statement}
This work uses existing scene-change benchmarks. Incorrect predictions
can lead to missed changes or unnecessary follow-up inspections.

\section*{Reproducibility Statement}
The implementation records reference camera assignments, synthetic events,
fitted scene models, effective-prior parameters, and per-image results.
All scene-specific parameters are frozen before target evaluation. The
appendices specify the observation transformations, statistical
approximations, update order, and effective-prior estimation used for the reported
method.

\bibliography{references}
\bibliographystyle{iclr2027_conference}

\appendix

\section{Image Evidence Processing and Gaussian Aggregation}
\label{app:frontend}

\subsection{Automatic reference editing and data generation}
\label{app:reference-edits}

The inputs are the fixed reference 3DGS and its registered images. Object
selection, edit parameters, labels, and filtering are determined automatically
from these inputs. We use SAM2.1 Hiera-Tiny automatic masks, images with a
maximum side length of 1008 pixels, and random seed 0. The following settings
are shared by all 20 scene instances.

\textbf{Reference camera assignment.}
We implement the approximately $5{:}1$ camera split described in
Section~\ref{sec:matched-data}. The unchanged-only subset contains
$\lceil N/6\rceil$ cameras for unchanged-response estimation, where $N$
is the total number of reference cameras. We select these cameras to
represent the scene's visible surfaces with a limited number of
observations. For camera $t$, let $\mathcal V_t$ denote its observable
reference Gaussians. We measure the distance between cameras $t$ and $l$
by the Jaccard distance
$1-|\mathcal V_t\cap\mathcal V_l|/|\mathcal V_t\cup\mathcal V_l|$,
so cameras observing similar surfaces are close. Greedy facility-location
selection adds cameras one at a time to minimise the total distance from
all reference cameras to their nearest selected camera. We then apply medoid swaps, replacing a selected camera with an unselected one whenever this reduces the total distance. Both steps require the editing subset to retain at least one available view of every synthetic object/change event. The editing subset provides labelled examples generated
by automatic 3DGS edits for learning response suppression, the
likelihood-ratio mapping, the effective pixel prior, and the contour
filter. Examples are included only where the corresponding edit
is available. With the image-evidence pipeline fixed, the unchanged-only
subset provides unchanged reference images for
estimating the Gaussian unchanged-response model and spatial co-response.
The two subsets contain different cameras but may observe overlapping
surfaces.

\textbf{Object proposals and cross-view association.}
We first generate candidate regions in each reference image using SAM2:
automatically placed point prompts produce masks, followed by filtering
of low-quality and duplicate results. To determine whether masks in
different views describe the same object, we associate each mask with
the reference Gaussians that mainly contribute to it.

For reference view $t$, let $a_{tug}$ be the rendering contribution
defined in Section~\ref{sec:preliminaries}.
At each pixel, we retain only the two largest contributions and set
the others to zero; denote these retained weights by $\widetilde a_{tug}$.
For a mask $M$, Gaussian $g$'s contributions inside and outside it are
\begin{equation}
A^{\mathrm{in}}_{tg}
=\sum_{u\in M}\widetilde a_{tug},
\qquad
A^{\mathrm{out}}_{tg}
=\sum_{u\in\Omega_t\setminus M}\widetilde a_{tug},
\label{eq:app-mask-contributions}
\end{equation}
where $\Omega_t$ is the reference image domain.
We assign $g$ to the mask when it contributes more to the inside than
the outside, allowing ties:
\begin{equation}
\mathcal S_t(M)=
\left\{g:
A^{\mathrm{in}}_{tg}>0,\quad
A^{\mathrm{in}}_{tg}\geq A^{\mathrm{out}}_{tg}
\right\}.
\label{eq:app-mask-support}
\end{equation}
This \emph{support set} therefore contains Gaussians with at least half
of their retained contribution inside the mask. Masks with empty support
sets are discarded.

To exploit view continuity in sequentially captured reference images,
we match each view with the next $K$ views in the reference sequence.
This reduces pairwise comparisons while allowing matches to be linked
through intermediate views into longer tracks. We use $K=2$.
We match masks across views by comparing the Gaussians assigned to them.
For each candidate pair, we use a hypergeometric test to determine
whether the two masks share more Gaussians than expected by chance.
The test accounts for both support-set sizes and uses all Gaussians
visible in either view, $\mathcal V_t\cup\mathcal V_l$, as the population. We apply the Benjamini--Hochberg procedure \citep{benjamini1995controlling} at level $0.01$
to filter chance overlaps across the candidate pairs.
Among the surviving pairs, we rank matches by Jaccard overlap
(intersection size divided by union size), breaking ties by the number
of shared Gaussians. We retain a pair only if each mask is the other's
highest-ranked match in the corresponding view. Finally, we link these matches into cross-view tracks, where each track is a
collection of masks associated with the same candidate object across
reference views, with at most one mask per view.

\textbf{Membership and edit-view validation.}
We estimate how strongly each editable Gaussian belongs to the candidate
object using the masks in its representative track. We use all available
views of this track, adding views from strongly linked tracks if fewer
than two are available. For each editable Gaussian $g$, we sum its
rendering contributions inside these masks and divide by its total
contribution across the selected views. This ratio defines its soft
membership $m_g\in[0,1]$: larger values indicate that more of its
contribution falls inside the object masks. Gaussians unobserved in
these views receive $m_g=0$.

Rendering these membership values produces a 2D object-support map in
each selected view. We convert the maps into binary masks using a shared
threshold chosen to maximise mean IoU with the corresponding SAM2 masks.
To limit search cost, we divide the quantile range $[0,1]$ into 256
equal intervals for support values above $10^{-7}$. The boundary
quantiles serve as candidate thresholds; we select the one maximising
mean IoU with the SAM2 masks, breaking ties in favour of larger thresholds.

To check consistency across views, we repeat this procedure while
leaving out one view at a time: both membership and the threshold are
fitted using the remaining views, then used to predict the object mask
in the omitted view. We retain candidates with mean IoU at least $0.65$
and minimum IoU at least $0.5$ against the omitted SAM2 masks.
For accepted candidates, we refit membership and the threshold using
all selected views. \textbf{All validation uses automatically generated
reference masks and requires no manually annotated change masks.}
\textbf{Gaussian recolouring.}
We compute a channelwise median RGB colour in the first selected view
over object Gaussians with membership weight at least $0.5$, falling back
to all selected object Gaussians if none meet this criterion. The target maximizes Euclidean RGB distance from this
median over the palette
$(0.90,0.15,0.18)$, $(0.12,0.72,0.24)$, $(0.12,0.35,0.92)$,
$(0.92,0.72,0.10)$, $(0.82,0.16,0.82)$, and $(0.08,0.76,0.82)$.
For each selected Gaussian, we interpolate its view-dependent RGB colour
towards this shared target with coefficient $0.45$ times its object
membership weight. The target is fixed across the object's rendered views.

\textbf{Gaussian removal and translation.}
Removal replaces Gaussian $g$'s opacity parameter $o_g$ by $(1-m_g)o_g$ for every selected Gaussian.
Translation moves Gaussians with $m_g\geq0.5$ by a common displacement,
leaving their other attributes fixed. Its magnitude is relative to object
extent. The search uses core Gaussians supported by at least half of the
anchor-track masks, intersected with the moved set; it falls back to the
moved set if that intersection is empty. The centre is their coordinatewise
median, and directional extents are measured between the 1st and 99th
percentiles, using extrema when fewer than eight points are available.

We select up to four evenly spaced seed views. In each seed view, eight
equally spaced directions, with a seeded random angular offset, lie in the
plane perpendicular to the centre's viewing direction. The initial
translation magnitude is $1.5$ times the object's 3D extent along the
candidate direction. Up to six scaling steps increase it to obtain a
projected centre displacement of $1.5$ projected object extents in the seed
view, corresponding to an extent-relative clearance parameter of $0.5$.
A view passes the geometric check when both projected centres and at least
two core points before and after translation are valid, some moved core points remain in
frame, the old--new projected interval overlap is at most $0.01$, and the
centre displacement is at least $1.5\times0.995$ original projected extents.
Interval overlap is divided by the shorter interval length.

Within each seed, valid directions are ranked by retained in-frame fraction,
image-border margin, smaller overlap, and shorter translation, in that order.
Each seed's selected displacement is then checked across all supporting
views. The final choice maximizes the number of valid views, followed by the
same ranking criteria using the worst valid view. At least one valid view
is required. A final rendering check requires a nonempty moved mask and
intersection with the original mask below $1\%$ of the smaller mask area.
Only passing views produce translation examples; an object without a valid
translation may still provide recolouring and removal examples.

\textbf{Changed, unchanged, and ignored regions.}
Let $M$ denote the candidate changed region: the reference SAM2 object mask
for recolouring and removal, or its union with the thresholded moved
Gaussian-support map for translation. Let $D$ be the channel-mean absolute
RGB difference between edited and original renders. Sorting $D$ in descending
order defines an effect footprint $E$ containing at least $99\%$ of its
summed mass, including ties at the cutoff. This footprint identifies edit
effects extending beyond the object mask.

Removal also checks the newly exposed background. For each Gaussian,
background reliability is the fraction of its observed contribution outside
the object's masks across the selected reference views. Rendering these
reliabilities after removal and dividing by remaining coverage gives a
background-provenance map. Pixels with remaining coverage below $0.05$ or
provenance below $0.5$ are invalid. For removal and translation, let $U$
contain such pixels inside the original object mask; for recolouring, $U$
is empty. The changed region is
$M\setminus U$, and the ignored region is $U\cup(E\setminus M)$.
All other pixels are labelled unchanged. Changed and ignored regions are disjoint.

We also store soft edited-Gaussian footprint maps. For recolouring and
removal this is the rendered membership; for translation it is
$1-(1-f_{\mathrm{old}})(1-f_{\mathrm{new}})$, where the two $f$ maps render
membership of the moved Gaussian set at its original and translated
positions. The hard masks, ignore masks, and soft maps remain associated
with the same object, camera, and edit.

\textbf{Filtering individual edits.}
At the generated image resolution, an original-lighting edit is retained
only if all conditions below hold. A five-pixel boundary neighbourhood is
formed by dilating the change mask with an $11\times11$ elliptical kernel;
ignored pixels outside this dilation are called distant ignored pixels.
\begin{center}
\small
\begin{tabular}{@{}p{0.69\linewidth}p{0.23\linewidth}@{}}
\toprule
Quantity & Acceptance condition \\
\midrule
Changed pixel count & $>0$ \\
Ignored area / image area & $\leq0.03$ \\
Ignored area / changed area & $\leq20$ \\
Distant ignored area / changed area & $\leq1$ \\
RGB edit-effect mass outside $M$ / total effect mass & $\leq0.20$ \\
Invalid background area / original object-mask area (removal) & $\leq0.50$ \\
\bottomrule
\end{tabular}
\end{center}
The effect-mass check uses $M$ before invalid interior pixels are excluded.
Translation requires at least one retained view. The filters operate on
reference-derived masks and render differences.

\textbf{Reference-fitted lighting variations.}
We fit a colour transform to explain differences between unchanged reference
images and their renders. Its affine component maps the view-dependent Gaussian colour
$\mathbf c_{tg}$ to
$\boldsymbol\gamma\odot\mathbf c_{tg}+\boldsymbol\beta$, with three channel gains
$\boldsymbol\gamma$ and offsets $\boldsymbol\beta$; $\odot$ denotes
componentwise multiplication. An optional directional component adds
$(\mathbf d^\top\mathbf n_g)\mathbf c_{tg}$, where $\mathbf d\in\mathbb R^3$ is
fitted and $\mathbf n_g$ is the Gaussian's smallest-axis direction, oriented
towards the mean reference camera position. Transformed colours are
restricted to $[0,1]$ before rendering.

We select whether to include the directional component using six reference
views, sampled at evenly spaced indices in the reference camera list.
For each view, both models are fitted at a maximum side length of 320 pixels
on even-parity checkerboard pixels with coverage above $0.05$, excluding
reliably tracked objects. Fewer than 64 fitting pixels triggers a fallback
to all covered even-parity pixels. Four iteratively reweighted least-squares
steps use per-pixel RGB residual norms $r$, scale
$s=\max(\operatorname{median}r,10^{-4})/0.6745$, and weights
$\min(1,1.345s/r)$. A ridge term towards the identity transform has strength
$\epsilon_{\mathrm{fp32}}\max(\operatorname{tr}(X^\top W X),1)$, where $X$
is the rendered colour-basis design matrix, $W$ contains fitting weights,
and $\epsilon_{\mathrm{fp32}}=2^{-23}$ is the machine epsilon for float32.
For model selection, predictions are evaluated on odd-parity covered object
pixels at the original resolution, falling back to non-object pixels when
fewer than 16 object pixels are available. The comparison averages squared RGB errors
capped at their per-view 90th percentile. We enable the directional component
when its mean error across the six views does not exceed the affine model's.
This is a fixed-size model-selection check; the selected views are not
chosen to optimize spatial coverage. It enables the directional component
for 7 of the 20 evaluated scene instances.

After model selection, coefficients are fitted separately for every reference
camera represented in the generated event list (43--108 cameras per instance
in our data). Thus, the six-view check does not limit the views used for
coefficient fitting or event generation. We retain three representative
coefficient vectors: the first is closest to the componentwise median,
followed by farthest-point selection after scaling by median absolute
deviation, with each scale bounded below by $10^{-4}$. For each camera, a fixed seeded choice selects one of the two
representatives farthest from its own fitted coefficients. That transform
is applied to both edited and unedited Gaussians, producing a relighting
condition alongside the identity condition. Relighting preserves the parent
change masks and soft targets; additional edit effects outside the changed regions
are included in the ignore mask. All lighting fitting and selection use
reference data before deployment.

\textbf{Paired-condition and object-level filtering.}
Relighting is applied only to edits that pass the individual-edit filters.
Its change masks and soft targets are inherited; the ignore mask also
includes any new edit-effect footprint outside the changed regions. We measure
this footprint from the relit edited--unedited rendering pair, so lighting
itself does not define changed pixels. A relit sample requires nonempty
changed regions and at most $20\%$ of edit-effect mass outside the union of changed
and ignored regions. We then retain an object--camera--edit tuple only when
both identity and relighting records are present exactly once, and both
have distant ignored area no larger than changed area. A failed condition
removes the pair.

Among objects with surviving pairs, more than half of the original
object--camera--edit records must have passed the individual-edit filters,
with at least two passing edit types. This fraction is computed before
relighting expansion. Finally, two candidates are linked as duplicates if
their anchor Gaussian sets have intersection over the smaller set at least
$0.9$, or Jaccard overlap at least $0.6$. We retain one candidate per connected
duplicate group, ranked by passing fraction, passing-record count, and the
number of Gaussians with membership at least $0.5$, with the object identifier
breaking any remaining tie. Only then are retained records assigned to the
reference camera subsets described above.

\textbf{Synthetic training observations.}
We transfer rendered interventions to the corresponding real reference
image. For reference image $I_t^{\mathrm{ref}}$, original rendering
$\widehat I_t$, and edited rendering $\widehat I_t^{\mathrm{edit}}$, the
synthetic observation is
$I_t^{\mathrm{syn}}=I_t^{\mathrm{ref}}+
\widehat I_t^{\mathrm{edit}}-\widehat I_t$, restricted to $[0,1]$.
This retains the reference image's image--render discrepancies while adding
the rendered edit. An unchanged control uses the unedited rendering under
the same lighting condition in place of $\widehat I_t^{\mathrm{edit}}$.
The image-evidence pipeline compares each resulting observation with the
original reference rendering. Applying the camera assignment above gives
11,384 edited training records across the 20 instances: 5,692 identity
and 5,692 relighting records. These are object--view--edit observations,
not distinct camera views.

\begin{figure}[t]
\centering
\includegraphics[width=\textwidth]{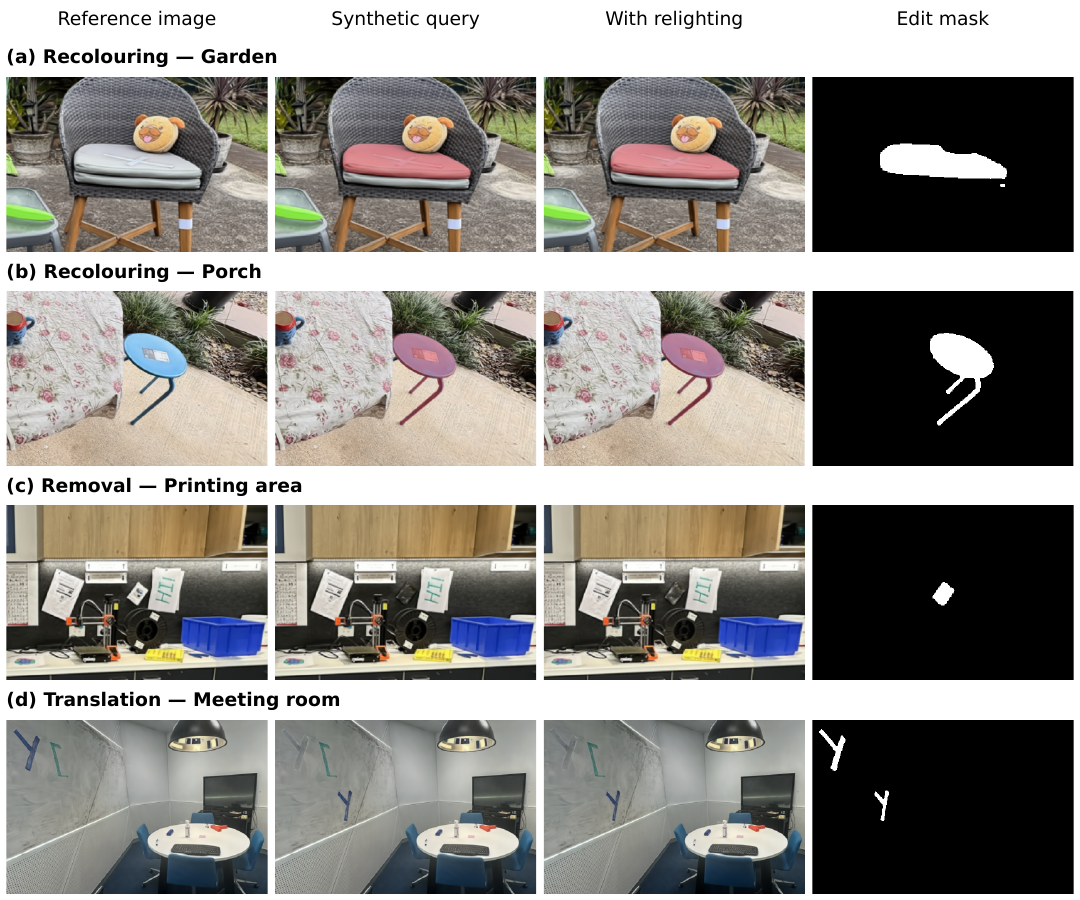}
\caption{Selected examples from the editing subset, showing recolouring, removal, and translation in four scenes. Each row uses the same crop across the real reference image, synthetic query, relit synthetic query, and edit mask. White marks regions labelled changed; translation includes both original and displaced locations. Relighting preserves these labels. Condition-specific ignore masks are omitted.}
\label{fig:reference-edit-examples}
\end{figure}

\subsection{Image evidence processing and aggregation}
\label{app:image-evidence}

\textbf{Query relocalisation.}
We follow O-SCD \citep{galappaththige2026oscd} to construct a database of
registered reference keyframes. XFeat keypoints \citep{potje2024xfeat} are matched across reference
images and triangulated with their known poses to obtain 3D points.
For each query, reference keyframes are selected by descriptor-match count,
and fundamental-matrix RANSAC rejects inconsistent matches.
Query matches to reference keypoints with known 3D positions establish
2D--3D correspondences. PnP-RANSAC estimates the query pose, followed by
reprojection-error refinement with camera intrinsics and reference points
held fixed. The fixed reference 3DGS is then rendered at the estimated pose.

\textbf{Input response and renderer-conditioned response suppression.}
The initial response uses the photometric and feature cues of
O-SCD \citep{galappaththige2026oscd}, rescaling their sum to $[0,1]$:
\begin{equation}
R_t^0(u)=\tfrac12\bigl[\widetilde D_t^{\mathrm{rgb}}(u)
                          +\widetilde D_t^{\mathrm{sem}}(u)\bigr].
\label{eq:initial-response}
\end{equation}
The photometric cue is
$D_t^{\mathrm{rgb}}(u)=(1-\lambda_{\mathrm{rgb}})L_{1,t}(u)+\lambda_{\mathrm{rgb}}L_{\mathrm{D\text{-}SSIM},t}(u)$,
where $L_{1,t}$ is the mean absolute RGB difference,
$L_{\mathrm{D\text{-}SSIM},t}=1-\operatorname{SSIM}_t$, and $\lambda_{\mathrm{rgb}}=0.2$.
For encoder $\phi$ with $K$ channels, the semantic discrepancy at feature-grid
location $\xi$ is
$D_t^{\mathrm{sem}}(\xi)=K^{-1}\|\phi(I_t)_\xi-\phi(\widehat I_t)_\xi\|_1$.
The SAM2 encoder is frozen, and both images are resized to $1024\times1024$
before feature extraction, following the O-SCD input construction.
RGB values lie in $[0,1]$. SSIM is channel-averaged over an $11\times11$
window \citep{wang2004ssim}, and $1/K$ averages the feature-channel
absolute differences. The tildes in Eq.~\ref{eq:initial-response} denote
min--max normalization for RGB, and min--max normalization followed by
bilinear upsampling for semantic differences. The semantic cue is then
rescaled with the RGB extrema and restricted to $[0,1]$.

Renderer-conditioned response suppression uses two logistic models with standardized
inputs. Their outputs $\ell_{\mathrm{base}}(x)$ and
$\ell_{\mathrm{full}}(x,z)$ are predicted change log odds. The vector $x$
contains the fused response, its $21\times21$ mean, $7\times7$ standard
deviation, and gradient magnitude. The full model additionally uses renderer
features $z$: coverage, gradients, ray-composition statistics, depth
dispersion, and response--renderer interactions. The correction rule
$\mathcal T(\ell_{\mathrm{base}},\ell_{\mathrm{full}})
=\min\{\ell_{\mathrm{full}}-\ell_{\mathrm{base}},0\}$ acts on response log odds:
\begin{equation}
R_t=\sigma\!\left(\operatorname{logit}R_t^0+
\mathcal T\bigl(\ell_{\mathrm{base}}(x_t),\ell_{\mathrm{full}}(x_t,z_t)\bigr)\right).
\label{eq:app-nuisance}
\end{equation}
Both regressions use L2 regularization. The correction is lower-bounded
by the 1st percentile of training logit differences and upper-bounded by
zero; it is disabled where rendered coverage is at most $10^{-3}$.
Fitting and deployment use this same corrected $R_t$.

\textbf{Response-to-evidence mapping.}
The mapping $\ell_{\mathrm{img}}$ is a monotone log likelihood ratio.
Following Section~\ref{sec:current}, we separately normalize response
histograms for changed regions of edited images $\mathcal H_1$, unchanged regions
of those images $\mathcal H_0^{\mathrm{edit}}$, and unedited reference controls $\mathcal H_0^{\mathrm{ref}}$.
Changed regions from all edit types contribute to $\mathcal H_1$.
The unchanged likelihood is their fixed mixture
$p_0=\tfrac12p_0^{\mathrm{edit}}+\tfrac12p_0^{\mathrm{ref}}$. These coefficients specify the
relative contributions of unchanged regions in edited images and unedited controls.

Isotonic regression \citep{zadrozny2002calibration} of the class-balanced
changed fraction, weighted by pooled histogram mass, determines contiguous
response blocks. For source $s\in\{1,\mathrm{edit},\mathrm{ref}\}$ and block $\mathcal B_b$,
let $M_{sb}$ be its normalized histogram mass and $N_s$ its effective
training-record count. Smoothed block probabilities and
the corresponding log likelihood ratio are
\begin{equation}
\widehat p_{sb}=
\frac{N_sM_{sb}+\tfrac12}{N_s+\tfrac12 K_{\mathcal B}},
\qquad
\ell_b=
\log\frac{\widehat p_{1b}}
{\tfrac12\widehat p_{\mathrm{edit},b}+\tfrac12\widehat p_{\mathrm{ref},b}},
\label{eq:app-table}
\end{equation}
where $K_{\mathcal B}$ is the number of current blocks. Effective counts
respect training-record grouping: object--camera units for edited-image classes
and camera units for unchanged controls. Violations of monotonicity after
smoothing are merged and re-estimated. We interpolate the block values onto
a 4,096-entry lookup table and keep the tails constant.
For a response block $b$, define
$\widehat p_{0b}=\tfrac12\widehat p_{\mathrm{edit},b}+\tfrac12\widehat p_{\mathrm{ref},b}$.
With equal prior weights for changed and unchanged regions, Bayes' rule gives
\begin{equation}
P_{\mathrm{emp},b}
=\frac{\widehat p_{1b}}{\widehat p_{1b}+\widehat p_{0b}}
=\sigma(\ell_b).
\label{eq:app-empirical-posterior}
\end{equation}
Applying the interpolated mapping gives $P_t=\sigma(\ell_{\mathrm{img}}(R_t))$.
These equal prior weights define the image-evidence scale. The scene-specific
effective pixel prior $\pi$ is fitted separately using the causal prediction
and field update in Eqs.~\ref{eq:final-support} and~\ref{eq:field-update}.

\textbf{Gaussian response moments.}
Both aggregation and support rendering use the rasterizer contributions
defined in Section~\ref{sec:preliminaries}. Zero denominators are guarded
numerically. Pixels with insufficient reference coverage receive zero change probability but remain included in evaluation.
For the second aggregation, define
\begin{equation}
r_{tg}=[\mathcal A_t(C_t)]_g,\qquad
v_{tg}=[\mathcal A_t(C_t^2)]_g-r_{tg}^2.
\label{eq:app-moments}
\end{equation}
Thus $v_{tg}$ measures dispersion across the current image footprint.
It is not a variance of the estimated mean across independent images.
The first aggregation applies the same operator to the current image
evidence $P_t$.

\textbf{Observation indicators.}
Neighbourhood evidence and history accumulation use binary support indicators,
\begin{equation}
\mathbb I_{tg}=\mathbb 1\{s_{tg}\ge 1\},
\label{eq:app-weight}
\end{equation}
where $s_{tg}$ is the rendering contribution summed over the image
(Section~\ref{sec:online-fusion}). The same rule is used in synthetic
sequence replay for effective-prior estimation and in online inference.
Every sufficiently supported Gaussian contributes one observation.

Unchanged-response fitting applies Eq.~\ref{eq:app-weight} to reference
images in the unchanged-only subset.
Thus each sufficiently supported observation receives equal fitting weight;
observations below the support threshold are excluded.
The resulting unchanged-response model and spatial co-response
coefficient $c$ are frozen before effective-prior estimation and
online inference.

\textbf{Reference neighbourhoods.}
We first retrieve eight nearest Gaussian centres, make the candidate edges
undirected, and remove duplicates and self-edges. Geometry and colour
affinities then filter and weight these connections. The final undirected
degree can exceed eight. For centre displacement $\Delta x_{gj}$, spatial
covariances $\Sigma_g,\Sigma_j$, and reference RGB colours $\mathbf k_g,\mathbf k_j$
from the zeroth-order spherical harmonic coefficients, the affinity is
\begin{equation}
\eta_{gj}=\exp\!\left[-\frac12
\Delta x_{gj}^{\mathsf T}(\Sigma_g+\Sigma_j)^{-1}\Delta x_{gj}
-\frac{\|\mathbf k_g-\mathbf k_j\|^2}{2s_{\mathrm{colour}}^2}\right].
\label{eq:app-affinity}
\end{equation}
The colour scale $s_{\mathrm{colour}}$ is the median nonzero colour distance among geometrically
overlapping candidates, floored at $0.001$. The covariance solve is
regularized numerically, and the geometric squared distance is capped at 50. Edges with combined
affinity no greater than $10^{-4}$ are removed.
The spatial covariance matrices describe Gaussian shape. The resulting
connections define each Gaussian's local reference neighbourhood.

\section{Unchanged-Scene Response Statistics}
\label{app:normal}

The frozen image-evidence pipeline is applied to the smaller camera group's
unchanged images. For image $t$, aggregating current evidence
$C_t$ gives the Gaussian response $r_{tg}=[\mathcal A_t(C_t)]_g$, with
observation indicator $\mathbb I_{tg}$ from Eq.~\ref{eq:app-weight}
in Appendix~\ref{app:frontend}.
All subsequent unchanged-response statistics refer to this normalized response
coordinate, including current-view Gaussian-aggregated support.

\textbf{Unchanged-response mean and uncertainty.}
All sums below use reference observations from the unchanged-only subset.
The scene mean is
$\mu_{\mathrm{scene}}=(\sum_{t,g}\mathbb I_{tg}r_{tg})/(\sum_{t,g}\mathbb I_{tg})$,
clipped to $[10^{-3},1-10^{-3}]$.
To derive the response moments, start with a
$\operatorname{Beta}(\kappa\mu_{\mathrm{scene}},
\kappa(1-\mu_{\mathrm{scene}}))$ prior for response level $\theta_g$,
with $\kappa>0$. Updating it with fractional response counts gives
\begin{equation}
\alpha_g=\kappa\mu_{\mathrm{scene}}+\sum_t \mathbb I_{tg}r_{tg},\qquad
\beta_g=\kappa(1-\mu_{\mathrm{scene}})+\sum_t \mathbb I_{tg}(1-r_{tg}).
\label{eq:app-beta}
\end{equation}
This follows from multiplying the prior density by
$\prod_t[\theta_g^{r_{tg}}(1-\theta_g)^{1-r_{tg}}]^{\mathbb I_{tg}}$
using Beta conjugacy \citep{gelman2013bayesiandataanalysis}; continuous
responses make this a soft-count approximation.
We use the limit $\kappa\to0^+$ of the mean and variance,
$\alpha_g/(\alpha_g+\beta_g)$ and
$\alpha_g\beta_g/[(\alpha_g+\beta_g)^2(\alpha_g+\beta_g+1)]$:
\begin{equation}
\mu_g=\begin{cases}
\displaystyle\frac{\sum_t \mathbb I_{tg}r_{tg}}{\sum_t \mathbb I_{tg}},&\sum_t \mathbb I_{tg}>0,\\
\mu_{\mathrm{scene}},&\sum_t \mathbb I_{tg}=0,
\end{cases}
\qquad
\tau_g^2=\frac{\mu_g(1-\mu_g)}{1+\sum_t \mathbb I_{tg}}.
\label{eq:app-beta-moments}
\end{equation}
Thus observed Gaussians use their own mean without shrinkage, while
unobserved Gaussians retain the limiting scene-based moments.
The implementation evaluates these moments directly, including zero
limiting variance when all observed responses are zero or one.
The variance represents uncertainty in this soft-count construction,
not the empirical predictive variance of a new image.
At inference, we use the approximate standardized coordinate
\begin{equation}
z_{tg}=
\frac{r_{tg}-\mu_g}{\sqrt{\tau_g^2+v_{tg}}}.
\label{eq:app-standardization}
\end{equation}
The denominator combines uncertainty in the fitted unchanged-response level
with dispersion within the current image footprint to define a working
standardization scale.

\textbf{Spatial co-response in unchanged scenes.}
To estimate spatial correlation, first standardize reference responses by
each Gaussian's empirical weighted mean and across-view standard deviation,
giving $\widetilde z_{tg}$. These empirical statistics are used for estimating
$c$, and are distinct from Eq.~\ref{eq:app-standardization}.
For neighbourhood affinity $\eta_{gj}$ and edge weight
$\omega^{\mathrm{fit}}_{tgj}=\eta_{gj}\min(\mathbb I_{tg},\mathbb I_{tj})$, summing over the unchanged-only subset gives
\begin{equation}
c=
\frac{\sum_{t,(g,j)} \omega^{\mathrm{fit}}_{tgj}\widetilde z_{tg}\widetilde z_{tj}}
     {\sum_{t,(g,j)} \omega^{\mathrm{fit}}_{tgj}}.
\label{eq:app-spatial-correlation}
\end{equation}

\section{Conditional Neighbourhood Bayes Factor}
\label{app:conditional}

\textbf{Weighted neighbourhood statistics.}
For neighbours $j$ of Gaussian $g$, let
$\omega_{tgj}=\eta_{gj}\mathbb I_{tj}$ and $W_{tg}=\sum_j \omega_{tgj}$. For $W_{tg}>0$,
\begin{equation}
\bar z_{tg}=\frac{\sum_j \omega_{tgj}z_{tj}}{W_{tg}},\qquad
d_{tg}=\frac{\sum_j \omega_{tgj}(z_{tj}-\bar z_{tg})^2}{W_{tg}},\qquad
n_{tg}=\frac{W_{tg}^2}{\sum_j \omega_{tgj}^2}.
\label{eq:app-neighbour-statistics}
\end{equation}
Here $\bar z_{tg}$ is the weighted mean, $d_{tg}$ the weighted dispersion,
and $n_{tg}$ the effective neighbour count. The working variance $V_{tg}$
of that mean accounts for dispersion and correlation; $\nu_{tg}$ is the
conditional residual variance used by the Bayes factor:
\begin{equation}
V_{tg}=
\frac{\max(1,d_{tg})}{\max(1,n_{tg})}
\left[1+c(n_{tg}-1)_+\right],\qquad
\nu_{tg}=\max(V_{tg}-c^2,\epsilon_\nu).
\label{eq:app-neighbour-variance}
\end{equation}
Here $(x)_+=\max(x,0)$, and we use variance floor $\epsilon_\nu=10^{-4}$.

\textbf{Why subtract the anchor response?}
Suppressing the frame and anchor indices, write $\bar z=\bar z_{tg}$,
$z_g=z_{tg}$, $V=V_{tg}$, and $\nu=\nu_{tg}$.
We use a zero-mean Gaussian working model under unchanged conditions.
Setting $\operatorname{Var}(z_g\mid \mathcal H_0)=1$ and approximating
$\operatorname{Cov}(\bar z,z_g\mid \mathcal H_0)$ by $c$ gives the conditional mean
\begin{equation}
\mathbb E[\bar z\mid z_g,\mathcal H_0]
=\frac{\operatorname{Cov}(\bar z,z_g\mid \mathcal H_0)}
{\operatorname{Var}(z_g\mid \mathcal H_0)}z_g\approx cz_g.
\label{eq:app-conditional-mean}
\end{equation}
The corresponding conditional variance is $V-c^2$.
Unit anchor variance is a fixed modelling
scale: Eq.~\ref{eq:app-standardization} does not guarantee that the
empirical variance of $z_g$ equals one. Using $c$ transfers the spatial
co-response estimated in Eq.~\ref{eq:app-spatial-correlation} to this
working model as a covariance approximation. The residual variance remains
neighbourhood-dependent through Eq.~\ref{eq:app-neighbour-variance}.
A shared increase $\delta$ in the responses of the anchor and neighbours leaves
residual mean $(1-c)\delta$. This motivates
$e=\bar z-cz_g$ and loading $a=1-c$.

Within this residual model, $\mathcal H_0$ denotes zero residual mean and $\mathcal H_1$ a shared
increase in response, with half-normal prior scale $\tau$:
\begin{equation}
\mathcal H_0:\ e\sim\mathcal N(0,\nu),\qquad
\mathcal H_1:\ e\mid\delta\sim\mathcal N(a\delta,\nu),\qquad
\delta\sim\operatorname{HalfNormal}(\tau),\quad \tau=1.
\label{eq:app-residual-model}
\end{equation}
The variance floor in Eq.~\ref{eq:app-neighbour-variance} also regularizes
cases where the plug-in covariance approximation is not positive definite.

\textbf{Integrating over the response increase.}
We evaluate the marginal likelihood of the residual $e$ with $a$ and
$\nu>0$ held fixed. For $\delta\geq0$, the half-normal density $p_+(\delta)$ is
\[
p_+(\delta)=2\mathcal N(\delta;0,\tau^2)
=\frac{2}{\sqrt{2\pi}\tau}
\exp\!\left(-\frac{\delta^2}{2\tau^2}\right).
\]
Here $\tau>0$ is the enhancement prior scale, fixed to $1$ in our method;
it is distinct from the fitted uncertainty $\tau_g^2$.
Since the enhancement $\delta$ is unknown, the marginal likelihood
$p(e\mid\mathcal H_1)$ integrates over this prior:
\[
p(e\mid\mathcal H_1)
=\int_0^\infty\mathcal N(e;a\delta,\nu)p_+(\delta)\,d\delta.
\]
The local Bayes factor $B$ therefore becomes
\[
B=\frac{p(e\mid\mathcal H_1)}{p(e\mid\mathcal H_0)}
=\int_0^\infty
\frac{\mathcal N(e;a\delta,\nu)}{\mathcal N(e;0,\nu)}
p_+(\delta)\,d\delta.
\]
Both likelihoods have variance $\nu$, so their normalization factors
cancel. Expanding the squared terms gives
\[
\begin{aligned}
\frac{\mathcal N(e;a\delta,\nu)}{\mathcal N(e;0,\nu)}
&=\exp\!\left(-\frac{(e-a\delta)^2}{2\nu}+\frac{e^2}{2\nu}\right)\\
&=\exp\!\left(-\frac{e^2-2ae\delta+a^2\delta^2-e^2}{2\nu}\right)\\
&=\exp\!\left(\frac{ae}{\nu}\delta-\frac{a^2}{2\nu}\delta^2\right).
\end{aligned}
\]
Multiplying by the half-normal prior yields
\[
B=\frac{2}{\sqrt{2\pi}\tau}
\int_0^\infty\exp\!\left[
\frac{ae}{\nu}\delta
-\frac12\left(\frac{a^2}{\nu}+\tau^{-2}\right)\delta^2
\right]d\delta.
\]
Define the coefficients $h=ae/\nu$ and $J=a^2/\nu+\tau^{-2}>0$.
Completing the square gives
\[
\begin{aligned}
h\delta-\frac12J\delta^2
&=-\frac J2\left(\delta^2-\frac{2h}{J}\delta\right)\\
&=-\frac J2\left[\left(\delta-\frac hJ\right)^2-\frac{h^2}{J^2}\right]\\
&=-\frac J2\left(\delta-\frac hJ\right)^2+\frac{h^2}{2J}.
\end{aligned}
\]
The last term does not depend on $\delta$, so
\[
B=\frac{2}{\sqrt{2\pi}\tau}
\exp\!\left(\frac{h^2}{2J}\right)
\int_0^\infty
\exp\!\left[-\frac J2\left(\delta-\frac hJ\right)^2\right]d\delta.
\]
To evaluate the remaining integral, substitute
$x=\sqrt J(\delta-h/J)$, giving $d\delta=dx/\sqrt J$.
The lower limit $\delta=0$ becomes $x=-h/\sqrt J$, while the upper
limit remains infinity. Thus
\[
B=\frac{2}{\tau\sqrt J}
\exp\!\left(\frac{h^2}{2J}\right)
\left[\frac{1}{\sqrt{2\pi}}
\int_{-h/\sqrt J}^{\infty}e^{-x^2/2}\,dx\right].
\]
Writing $\Phi$ for the standard normal cumulative distribution function,
the bracketed term equals
\[
1-\Phi\!\left(-\frac h{\sqrt J}\right)
=\Phi\!\left(\frac h{\sqrt J}\right),
\]
by symmetry of the standard normal distribution. The resulting Bayes
factor $B$ is
\begin{equation}
B=\frac{2}{\tau\sqrt J}
  \exp\!\left(\frac{h^2}{2J}\right)
  \Phi\!\left(\frac{h}{\sqrt J}\right).
\label{eq:app-bf-integral}
\end{equation}
Taking logarithms gives the log evidence $\log B$:
\begin{equation}
\log B=\log2-\log\tau-\tfrac12\log J+
       \frac{h^2}{2J}+\log\Phi\!\left(\frac{h}{\sqrt J}\right).
\label{eq:app-bf-closed}
\end{equation}
With $\tau=1$, restoring the indices gives
$h_{tg}=(1-c)e_{tg}/\nu_{tg}$ and $J_{tg}=(1-c)^2/\nu_{tg}+1$,
recovering Eq.~\ref{eq:conditional-evidence}.
The Gaussian likelihood and half-normal prior make the exponent quadratic
in $\delta$, reducing marginalization to a standard normal CDF evaluation.
The integral is therefore evaluated analytically under the specified
residual model, without numerical integration or iterative optimization.
If no neighbour has positive observation weight, we use neutral evidence
$B=1$, so the log evidence is zero.

\textbf{Implementation bounds.}
We evaluate $\log\Phi(h/\sqrt J)$ directly using the numerically stable
\texttt{log\_ndtr} function, avoiding an explicit CDF floor.
The implementation
floors $\tau_g^2+v_{tg}$ at $10^{-6}$ to avoid division by zero, while
$\epsilon_\nu=10^{-4}$ prevents a degenerate residual variance.

We clip standardized responses and per-observation log Bayes factors to
$[-12,12]$ to limit the influence of extreme observations. These bounds
are fixed across scenes.
The dispersion floor and correlation adjustment in
Eq.~\ref{eq:app-neighbour-variance} are part of the working model.

\section{Accumulated Support and Causal Decoding}
\label{app:history}

\textbf{Generalized pixel posterior.}
Let $y=1$ denote pixel change and $y=0$ no change. For a support value
$p\in(0,1)$, define the binary cross-entropy loss and the sum used for fusion:
\begin{equation}
\begin{aligned}
\operatorname{BCE}(y,p)&=-y\log p-(1-y)\log(1-p),\\
\mathcal L_t(y,u)&=\operatorname{BCE}(y,C_t(u))+
                   \operatorname{BCE}(y,H_t(u)).
\end{aligned}
\label{eq:app-composite-loss}
\end{equation}
The two terms assign unit weight to agreement with current evidence and
historical support. Let $q_t(y\mid u)$ denote the generalised posterior
distribution over the two labels at pixel $u$, formed from $C_t(u)$,
$H_t(u)$, and prior $\pi$. A loss-based Bernoulli update
\citep{bissiri2016general} gives
\begin{equation}
q_t(y\mid u)=
\frac{\pi^y(1-\pi)^{1-y}\exp[-\mathcal L_t(y,u)]}
{\sum_{v\in\{0,1\}}\pi^v(1-\pi)^{1-v}\exp[-\mathcal L_t(v,u)]}.
\label{eq:app-generalized-posterior}
\end{equation}
Taking the ratio of the two label probabilities gives
\begin{equation}
\frac{q_t(1\mid u)}{q_t(0\mid u)}
=\frac{\pi}{1-\pi}
 \frac{C_t(u)}{1-C_t(u)}
 \frac{H_t(u)}{1-H_t(u)}.
\label{eq:app-posterior-odds}
\end{equation}
The scalar map $Q_t^{\mathrm{pixel}}(u)=q_t(y=1\mid u)$ contains the generalised posterior
probability of change; $q_t(y=0\mid u)=1-Q_t^{\mathrm{pixel}}(u)$. This gives the odds update in
Eq.~\ref{eq:final-support}, equivalently
$Q_t^{\mathrm{pixel}}=\sigma(\operatorname{logit}\pi+\operatorname{logit}C_t+
\operatorname{logit}H_t)$. The implementation evaluates this closed
form; the loss representation adds no training step. It specifies a
composite evidence update rather than a factorization into independent
observations. Selecting $\pi$ by synthetic training IoU is a separate optimization
objective from $\mathcal L_t$; neither that selection nor normalization
guarantees exact probabilities on the target distribution.

\textbf{Accumulated neighbourhood evidence.}
After pixel prediction, each sufficiently supported observation adds its
log Bayes factor to the previous accumulated field $F_{t-1}$, giving the updated accumulated field $F_t$:
\begin{equation}
F_{tg}=F_{t-1,g}+\mathbb I_{tg}\log B_{tg}.
\label{eq:app-state}
\end{equation}
The accumulator starts at zero; the decoded support depends directly on $F_{tg}$.
Neither initial image responses nor final pixel posteriors are written into this
state. An observed Gaussian without usable neighbours contributes zero log
evidence and leaves the accumulator unchanged.

\textbf{Rendering the previous support.}
The previous accumulated field $F_{t-1}$ is decoded per Gaussian before rendering:
\begin{equation}
H_t(u)=\frac{\sum_g a_{tug}\sigma(F_{t-1,g})}{\chi_{tu}}.
\label{eq:app-render-history}
\end{equation}
The sigmoid is applied to each Gaussian before spatial averaging; only
then do we take $\operatorname{logit}H_t$ for the pixel update. The
normalized contributions $a_{tug}/\chi_{tu}$ mix support from reference
Gaussians associated with the pixel. With no previous observations,
$F=0$, giving $\sigma(F)=0.5$ and neutral history on covered pixels. The
effective prior $\pi$ is applied after this rendering and leaves the
Gaussian evidence accumulator unchanged.

\textbf{Effective-prior search parameterization and storage.}
Let $Z=\operatorname{logit}C+\operatorname{logit}H$.
The generalised posterior is $\sigma(Z+\operatorname{logit}\pi)=\sigma(Z-b_\pi)$,
where $b_\pi=-\operatorname{logit}\pi$. For threshold search the implementation
uses the equivalent coordinate $s=Z/2$, $\beta=b_\pi/2$, so
$\sigma(Z-b_\pi)=\sigma(2(s-\beta))$ and the mask is determined by
$s\geq\beta$. The scale factor is retained when evaluating scores.
The fitted prior is stored directly in log-odds form.

Before evaluating their logits, we clamp $C$ and $H$ to
$[\epsilon_{\mathrm{fp32}},1-\epsilon_{\mathrm{fp32}}]$, where
$\epsilon_{\mathrm{fp32}}=2^{-23}$ is the machine epsilon for float32,
to avoid infinities at 0 and 1.\footnote{See PyTorch's
\href{https://docs.pytorch.org/docs/main/generated/torch.distributions.utils.clamp_probs.html}{\texttt{clamp\_probs} documentation}.}
Synthetic replay and online detection
use the same floating-point probability maps, coverage mask, and decoding rule.
\textbf{Execution order.}
For each query image, the implementation:
\begin{enumerate}
  \item computes the image--render response, applies the learned response suppression,
  and applies the monotone evidence mapping;
  \item aggregates this score onto the Gaussians, renders Gaussian-aggregated
  support, and combines the two to obtain $C_t$;
  \item aggregates $C_t$ once, standardizes against the unchanged-response
  model, and computes conditional neighbourhood evidence;
  \item renders history from the previous accumulated field $F_{t-1}$ and combines it with $C_t$ using
  the fitted effective prior to obtain the pixel posterior;
  \item adds the current log Bayes factor to the evidence accumulator
  for each sufficiently supported Gaussian
  exactly once.
\end{enumerate}
Object-guided mask refinement acts on the image prediction without
modifying history. A new target sequence starts with empty state.

\paragraph{Insufficient reference coverage.}
The observation indicator $\mathbb I_{tg}$ prevents history updates when Gaussian $g$'s total rendering contribution is below one; without usable neighbours, $\log B_{tg}=0$.
Pixels with reference coverage at most $10^{-3}$ are assigned zero change probability and excluded from object completion, but remain in the evaluation domain.
These rules avoid unsupported updates but cannot recover missing geometry, so changes in uncovered regions may be missed.
The indicator measures rendering contribution, not geometric accuracy.

\section{Reliability-Weighted Effective-Prior Estimation}
\label{app:decision}

The effective pixel prior is fitted for the synthetic training objective in
Eq.~\ref{eq:decision-objective}; it specifies an operating point, not
an estimate of deployment change frequency. We implement this optimization
through the equivalent threshold coordinate $b_\pi=-\operatorname{logit}\pi$,
using $s=Z/2$ and $\beta=b_\pi/2$ for the deployment-matched search below.
This one-to-one reparameterization preserves the objective and its
optimality gap.
For image $i$, foreground IoU is
$\operatorname{IoU}(\widehat Y_i^{\mathrm{pixel}}(\pi),Y_i)
=\operatorname{TP}_i(\pi)/(N_i^++\operatorname{FP}_i(\pi))$,
where $N_i^+$ counts labelled changed pixels, and
$\operatorname{TP}_i$ and $\operatorname{FP}_i$ count true and false
positives before refinement.

\textbf{Comparing unchanged-response tails across events.}
For synthetic image $i$, let $H_{0,i}$ be its labelled unchanged pixels
and $s_i(x)$ the decision score at pixel $x$. For a score cutoff $v$,
the unchanged survival curve is the fraction of these pixels exceeding it:
\begin{equation}
T_i(v)=
\frac{|\{x\in H_{0,i}:s_i(x)\geq v\}|}{|H_{0,i}|}.
\label{eq:app-tail}
\end{equation}
An edit event is an object/edit-type combination. First average $T_i$ within
each event $e$ to obtain $\overline T_e$. For an image from event $e$,
its reference curve is the pointwise median of $\overline T_{e'}$ over
other events $e'\ne e$, denoted $T_{-e}^{\mathrm{ref}}$.
This forms an unchanged-response reference across events, limiting the
influence of repeated images of any one event.

Let $e(i)$ be image $i$'s event and $[v_{\min},v_{\max}]$ the range
of training-score cutoffs used to compare tails. Excess high response is
\begin{equation}
d_i=\frac{1}{v_{\max}-v_{\min}}
\int_{v_{\min}}^{v_{\max}}
\left[T_i(v)-T_{-e(i)}^{\mathrm{ref}}(v)\right]_+\,dv.
\label{eq:app-tail-distance}
\end{equation}
The implementation evaluates this integral by the trapezoidal rule on the
fixed training-score cutoffs. The subsequent decision-threshold search
operates over the full score range.

Let $D_e=\operatorname{median}_{i:e(i)=e}d_i$.
For event $e$, compute the centre $m_{-e}$ and scale $a_{-e}$ from the other
event medians:
\begin{equation}
\begin{aligned}
m_{-e}&=\operatorname{median}_{e'\ne e}D_{e'},&
a_{-e}&=1.4826\,\operatorname{median}_{e'\ne e}|D_{e'}-m_{-e}|,\\
z_i^{\mathrm{tail}}&=\left[\frac{d_i-m_{-e(i)}}{a_{-e(i)}}\right]_+,&
\rho_i&=\frac{1}{1+(z_i^{\mathrm{tail}})^2}.
\end{aligned}
\label{eq:app-reliability}
\end{equation}
Images with $z_i^{\mathrm{tail}}=0$ receive weight one; standardized excesses
of one and two receive weights $1/2$ and $1/5$.
The factor $1.4826$ scales the median absolute deviation (MAD) to match
the standard deviation under a Gaussian model.
The deployed fit requires positive across-event MAD. We retain the original
pixel counts and IoU definitions; $\rho_i$ weights each image's contribution
to the mean IoU according to its unchanged-response reliability.

\textbf{Training sequences and fixed weights.}
Sequences of images from the editing subset use the same image-evidence pipeline, unchanged-response model,
history update order, and pixel prediction rule as online inference. The weights in
Eq.~\ref{eq:app-reliability} are normalized as $\widetilde \rho_i = \rho_i / \sum_j \rho_j$, computed once before searching for
the threshold and remain fixed throughout optimization. Synthetic labels
supply the unchanged regions used for reliability weighting and the
changed regions used for IoU.

\textbf{A valid branch-and-bound upper bound.}
Let $J(v)$ be the reliability-weighted IoU at decision threshold $v$ in
the $s$ coordinate. Increasing $v$ can only decrease both TP and FP.
Consequently, for any interval $[l,r]$ and any $v$ within it,
\begin{equation}
J(v)\leq
U(l,r)=\sum_i\widetilde \rho_i\,
\frac{\operatorname{TP}_i(l)}
     {N_i^+ + \operatorname{FP}_i(r)}.
\label{eq:app-search-bound}
\end{equation}
The numerator at $l$ bounds candidate numerators from above, and the
denominator at $r$ bounds denominators from below. Nonnegative weights
preserve the bound under summation, even if no threshold attains it.

The search starts from a range enclosing the training scores, including
all-positive and all-negative decisions. Sorted changed and unchanged
scores support exact count queries by binary search. The optimizer
maintains the best evaluated objective $J_{\mathrm{best}}$, splits the
remaining interval with the highest bound, evaluates new candidates, and
prunes intervals whose bound cannot improve the best objective value found so far.
It stops when
\begin{equation}
\max_{\text{remaining intervals}}U(l,r)-J_{\mathrm{best}}
\leq 10^{-4}.
\label{eq:app-search-stop}
\end{equation}
This is $0.01$ percentage points in the synthetic training IoU objective.
Midpoint subdivision does not require IoU to be monotone or unimodal.
The certificate applies to the recorded, weighted training objective
with the deployment pixel prediction rule.

\section{Object-Guided Mask Refinement}
\label{app:completion}

Object-guided mask refinement uses shared SAM2 image embeddings to
suppress unsupported detections and complete retained object regions.

\textbf{Contour learning from pixel proposals.}
We process sequences of images from the editing subset with the
fixed pixel models, effective prior, and causal history. Thresholding the
source pixel posterior at $0.5$ produces connected components whose boxes
prompt SAM2 on the synthetic query and reference rendering. Predictions
supply prompts only: change masks from automatic reference edits
(Appendix~\ref{app:reference-edits}) supply pixel labels, and ignored pixels
are excluded.

The classifier uses 18 query/reference-symmetric features. We rank SAM2's
three candidates independently in each view by predicted IoU and pair them
by rank, forming paired mask logits, quality scores, and their per-side products. Each pair is replaced by its
maximum and minimum, using the same transformation during fitting and
inference. This retains support from either image regardless of which
side contains the object. Within each
source image, existing class-balanced pixel weights are multiplied once
by its fixed prior-fitting reliability $\rho_i$. Global balancing then gives
each class total weight $0.5$ within its fitting or validation split.
These weights are used for feature standardization, L2-regularized logistic
regression, and Platt sigmoid scaling \citep{platt2000probabilities}. At inference,
standardization and both fitted affine transformations are folded into a
single affine score $\ell_{\mathrm{contour}}$ on the unstandardized symmetric features. The contour fitting/validation
split is by edited object; upstream pixel models still use the larger
camera group. The pixel parameters and prior are unchanged.

\textbf{Online contour filtering.}
We decode masks on the query image and reference rendering using
bounding-box prompts from detected connected components. The fitted classifier converts their
mask-comparison features into $\ell_{\mathrm{contour}}(u)$. Its non-positive
part gives the contour-filtered score $Q_t^{\mathrm{contour}}(u)$ through
Eq.~\ref{eq:contour-refinement}.

\textbf{Object completion.}
Point prompts at response peaks of the surviving components produce SAM2
candidates in the query image. For each prompt, we select the candidate
with highest predicted mask quality. For a candidate mask $M$ over covered
pixels, its mean log-odds evidence is
$\overline L_M=|M|^{-1}\sum_{u\in M}\operatorname{logit}Q_t^{\mathrm{pixel}}(u)$,
where $|M|$ is the number of covered pixels in the mask. We retain masks
with $\overline L_M\geq0$ and combine their confidence
$\sigma(\overline L_M)$ with $Q_t^{\mathrm{contour}}$ to obtain the refined change
score $Q_t^{\mathrm{full}}$ using Eq.~\ref{eq:refined-output}.
The component filter skips regions smaller than the SAM feature
cell, and empty or uncovered masks are excluded. The pixel prior is
already included in $Q_t^{\mathrm{pixel}}$; refinement modifies only the final output.

\section{Additional Quantitative Results}
\label{app:additional-results}

Table~\ref{tab:per-scene} reports the full 20-instance comparison for
the pixel and full outputs in Table~\ref{tab:overall}.

\begin{table}[t]
\centering
\caption{PASLCD comparison using our evaluator over 20 scene-instances
(\%, higher is better). O-SCD(O) is online; O-SCD(R) includes offline
refinement. Ours (pixel) is the pixel output; Ours (full) adds object-guided
mask refinement. Our results average two complete runs; Total summarizes all 20 instances.
Cell backgrounds mark \textcolor{rankfirstbg}{\rule{0.8em}{0.8em}}~first,
\textcolor{ranksecondbg}{\rule{0.8em}{0.8em}}~second, and
\textcolor{rankthirdbg}{\rule{0.8em}{0.8em}}~third place for each row and
metric; these values are bold, and ties at the displayed precision share a rank.}
\label{tab:per-scene}
\small
\setlength{\tabcolsep}{3pt}
\renewcommand{\arraystretch}{1}
\begin{tabular*}{\textwidth}{@{\extracolsep{\fill}}lc*{8}{c}@{}}
\toprule
Scene & Inst. & \multicolumn{2}{c}{\shortstack{O-SCD(O)\\{\scriptsize CVPR 2026}}}
& \multicolumn{2}{c}{\shortstack{O-SCD(R)\\{\scriptsize CVPR 2026}}}
& \multicolumn{2}{c}{\shortstack{Ours\\pixel}}
& \multicolumn{2}{c}{\shortstack{Ours\\full}}\\
\cmidrule(lr){3-4}\cmidrule(lr){5-6}\cmidrule(lr){7-8}\cmidrule(lr){9-10}
& & mIoU & F1 & mIoU & F1 & mIoU & F1 & mIoU & F1\\
\midrule
Cantina & 1 & 50.24 & 65.86 & \cellcolor{ranksecondbg}\textbf{62.05} & \cellcolor{ranksecondbg}\textbf{76.05} & \cellcolor{rankthirdbg}\textbf{57.84} & \cellcolor{rankthirdbg}\textbf{72.68} & \cellcolor{rankfirstbg}\textbf{62.79} & \cellcolor{rankfirstbg}\textbf{76.23}\\
 & 2 & 49.38 & 65.16 & \cellcolor{rankthirdbg}\textbf{54.75} & \cellcolor{rankthirdbg}\textbf{70.19} & \cellcolor{ranksecondbg}\textbf{55.29} & \cellcolor{ranksecondbg}\textbf{70.48} & \cellcolor{rankfirstbg}\textbf{58.98} & \cellcolor{rankfirstbg}\textbf{73.16}\\
Garden & 1 & 49.58 & 65.52 & \cellcolor{rankthirdbg}\textbf{52.28} & \cellcolor{rankthirdbg}\textbf{67.07} & \cellcolor{ranksecondbg}\textbf{68.38} & \cellcolor{ranksecondbg}\textbf{80.43} & \cellcolor{rankfirstbg}\textbf{75.31} & \cellcolor{rankfirstbg}\textbf{85.33}\\
 & 2 & 51.54 & 67.78 & \cellcolor{rankthirdbg}\textbf{56.62} & \cellcolor{rankthirdbg}\textbf{72.27} & \cellcolor{ranksecondbg}\textbf{67.44} & \cellcolor{ranksecondbg}\textbf{80.37} & \cellcolor{rankfirstbg}\textbf{74.77} & \cellcolor{rankfirstbg}\textbf{85.51}\\
Lounge & 1 & 54.74 & 70.38 & \cellcolor{rankthirdbg}\textbf{60.04} & \cellcolor{rankthirdbg}\textbf{74.91} & \cellcolor{ranksecondbg}\textbf{68.96} & \cellcolor{ranksecondbg}\textbf{81.41} & \cellcolor{rankfirstbg}\textbf{75.18} & \cellcolor{rankfirstbg}\textbf{85.63}\\
 & 2 & 47.75 & \cellcolor{rankthirdbg}\textbf{64.12} & \cellcolor{rankthirdbg}\textbf{47.96} & 63.78 & \cellcolor{ranksecondbg}\textbf{48.74} & \cellcolor{ranksecondbg}\textbf{64.63} & \cellcolor{rankfirstbg}\textbf{59.28} & \cellcolor{rankfirstbg}\textbf{73.31}\\
Lunch room & 1 & 38.85 & 53.95 & \cellcolor{ranksecondbg}\textbf{49.19} & \cellcolor{rankfirstbg}\textbf{64.36} & \cellcolor{rankthirdbg}\textbf{48.53} & \cellcolor{rankthirdbg}\textbf{63.18} & \cellcolor{rankfirstbg}\textbf{49.83} & \cellcolor{ranksecondbg}\textbf{63.59}\\
 & 2 & 38.18 & 53.48 & \cellcolor{ranksecondbg}\textbf{49.15} & \cellcolor{rankfirstbg}\textbf{64.60} & \cellcolor{rankthirdbg}\textbf{47.61} & \cellcolor{rankthirdbg}\textbf{62.47} & \cellcolor{rankfirstbg}\textbf{51.14} & \cellcolor{ranksecondbg}\textbf{64.51}\\
Meeting room & 1 & 49.17 & 65.39 & \cellcolor{rankthirdbg}\textbf{55.44} & \cellcolor{rankthirdbg}\textbf{70.83} & \cellcolor{ranksecondbg}\textbf{57.28} & \cellcolor{ranksecondbg}\textbf{72.07} & \cellcolor{rankfirstbg}\textbf{62.92} & \cellcolor{rankfirstbg}\textbf{76.00}\\
 & 2 & 45.53 & 61.83 & \cellcolor{rankthirdbg}\textbf{51.66} & \cellcolor{rankthirdbg}\textbf{67.50} & \cellcolor{ranksecondbg}\textbf{53.07} & \cellcolor{ranksecondbg}\textbf{68.38} & \cellcolor{rankfirstbg}\textbf{64.14} & \cellcolor{rankfirstbg}\textbf{76.62}\\
Playground & 1 & \cellcolor{rankthirdbg}\textbf{38.74} & \cellcolor{rankthirdbg}\textbf{54.18} & 33.89 & 49.68 & \cellcolor{ranksecondbg}\textbf{51.89} & \cellcolor{ranksecondbg}\textbf{66.69} & \cellcolor{rankfirstbg}\textbf{56.47} & \cellcolor{rankfirstbg}\textbf{70.08}\\
 & 2 & \cellcolor{rankthirdbg}\textbf{36.22} & \cellcolor{rankthirdbg}\textbf{51.64} & 31.19 & 46.13 & \cellcolor{ranksecondbg}\textbf{51.20} & \cellcolor{ranksecondbg}\textbf{65.98} & \cellcolor{rankfirstbg}\textbf{57.28} & \cellcolor{rankfirstbg}\textbf{70.29}\\
Porch & 1 & 57.02 & 71.82 & \cellcolor{rankthirdbg}\textbf{71.48} & \cellcolor{rankthirdbg}\textbf{83.14} & \cellcolor{ranksecondbg}\textbf{73.63} & \cellcolor{ranksecondbg}\textbf{84.33} & \cellcolor{rankfirstbg}\textbf{81.40} & \cellcolor{rankfirstbg}\textbf{89.03}\\
 & 2 & 53.03 & 68.58 & \cellcolor{ranksecondbg}\textbf{63.60} & \cellcolor{ranksecondbg}\textbf{77.46} & \cellcolor{rankthirdbg}\textbf{63.30} & \cellcolor{rankthirdbg}\textbf{76.79} & \cellcolor{rankfirstbg}\textbf{70.83} & \cellcolor{rankfirstbg}\textbf{81.89}\\
Pots & 1 & 60.83 & 75.37 & \cellcolor{rankthirdbg}\textbf{65.20} & \cellcolor{rankthirdbg}\textbf{78.49} & \cellcolor{ranksecondbg}\textbf{73.48} & \cellcolor{ranksecondbg}\textbf{84.64} & \cellcolor{rankfirstbg}\textbf{87.11} & \cellcolor{rankfirstbg}\textbf{93.05}\\
 & 2 & \cellcolor{rankthirdbg}\textbf{57.67} & \cellcolor{rankthirdbg}\textbf{72.63} & \cellcolor{ranksecondbg}\textbf{63.04} & \cellcolor{ranksecondbg}\textbf{76.89} & 55.75 & 71.04 & \cellcolor{rankfirstbg}\textbf{74.84} & \cellcolor{rankfirstbg}\textbf{85.17}\\
Printing area & 1 & 60.69 & 74.60 & \cellcolor{rankthirdbg}\textbf{69.68} & \cellcolor{rankthirdbg}\textbf{81.82} & \cellcolor{ranksecondbg}\textbf{75.97} & \cellcolor{ranksecondbg}\textbf{86.11} & \cellcolor{rankfirstbg}\textbf{78.06} & \cellcolor{rankfirstbg}\textbf{87.46}\\
 & 2 & 52.11 & 67.94 & \cellcolor{rankthirdbg}\textbf{57.86} & \cellcolor{rankthirdbg}\textbf{73.03} & \cellcolor{ranksecondbg}\textbf{61.95} & \cellcolor{ranksecondbg}\textbf{76.24} & \cellcolor{rankfirstbg}\textbf{66.95} & \cellcolor{rankfirstbg}\textbf{79.82}\\
Zen & 1 & 47.42 & 60.72 & \cellcolor{rankthirdbg}\textbf{62.23} & \cellcolor{rankthirdbg}\textbf{72.33} & \cellcolor{ranksecondbg}\textbf{67.46} & \cellcolor{ranksecondbg}\textbf{76.15} & \cellcolor{rankfirstbg}\textbf{75.55} & \cellcolor{rankfirstbg}\textbf{81.31}\\
 & 2 & 46.17 & 60.97 & \cellcolor{ranksecondbg}\textbf{62.54} & \cellcolor{ranksecondbg}\textbf{74.88} & \cellcolor{rankthirdbg}\textbf{59.52} & \cellcolor{rankthirdbg}\textbf{71.92} & \cellcolor{rankfirstbg}\textbf{70.01} & \cellcolor{rankfirstbg}\textbf{79.47}\\
\midrule
\multicolumn{2}{l}{\textbf{Total}} & 49.24 & 64.60 & \cellcolor{rankthirdbg}\textbf{55.99} & \cellcolor{rankthirdbg}\textbf{70.27} & \cellcolor{ranksecondbg}\textbf{60.36} & \cellcolor{ranksecondbg}\textbf{73.80} & \cellcolor{rankfirstbg}\textbf{67.64} & \cellcolor{rankfirstbg}\textbf{78.87}\\
\bottomrule
\end{tabular*}
\end{table}

\subsection{Cross-instance Reference Reuse}
\label{app:cross-instance}

\begin{table}[t]
\centering
\caption{Same-scene cross-instance reference reuse on PASLCD (\%). Original pairing uses each instance's own reference; cross-instance pairing uses the other instance's reference and fitted parameters. Both average two runs of 20 sequences.}
\label{tab:cross-instance}
\small
\begin{tabular}{lrrrr}
\toprule
& \multicolumn{2}{c}{Original pairing} & \multicolumn{2}{c}{Cross-instance}\\
\cmidrule(lr){2-3}\cmidrule(lr){4-5}
Output & mIoU & F1 & mIoU & F1\\
\midrule
Ours (pixel) & 60.36 & 73.80 & 59.93 & 73.39\\
Ours (full) & 67.64 & 78.87 & 67.15 & 78.41\\
\bottomrule
\end{tabular}
\end{table}

Each PASLCD scene has two instances. We retain the reference reconstruction and all fitted parameters from one instance and evaluate queries from the other, in both directions. This gives 20 directed pairs, each with 25 frames. The reused parameters include response suppression, the LLR mapping, Gaussian-indexed unchanged-response statistics and spatial co-response, the pixel prior, and the contour filter. No parameters are refitted or selected using target data, and each sequence starts with empty history.

Queries are processed on the reference image canvas using reference-derived poses and initial difference cues. Query RGB is resized to this canvas for refinement, and the fixed reference model is rendered at the query pose. Predictions are evaluated against the target instance's original masks using the same resizing and thresholding convention as the main experiments. Table~\ref{tab:cross-instance-details} reports all pairs. The original and cross-instance aggregate comparisons each contain every target instance once. Because the paired query sets are highly similar, this experiment tests within-scene reference reuse rather than cross-scene generalization. Runtime was not measured.

\begin{table}[t]
\centering
\caption{Cross-instance results for all 20 directed PASLCD pairs (\%). Direction denotes reference instance $\to$ query instance. Pixel and full outputs use the same fixed reference reconstruction and fitted parameters.}
\label{tab:cross-instance-details}
\small
\begin{tabular*}{\textwidth}{@{\extracolsep{\fill}}lcrrrr@{}}
\toprule
& & \multicolumn{2}{c}{Pixel output} & \multicolumn{2}{c}{Full output}\\
\cmidrule(lr){3-4}\cmidrule(lr){5-6}
Scene & Direction & mIoU & F1 & mIoU & F1\\
\midrule
Cantina & $1\to 2$ & 57.82 & 72.61 & 62.87 & 76.17\\
Cantina & $2\to 1$ & 54.89 & 70.08 & 57.88 & 72.29\\
Garden & $1\to 2$ & 67.21 & 79.65 & 73.76 & 84.35\\
Garden & $2\to 1$ & 66.99 & 80.03 & 74.53 & 85.33\\
Lounge & $1\to 2$ & 68.94 & 81.48 & 74.88 & 85.47\\
Lounge & $2\to 1$ & 48.10 & 64.06 & 58.40 & 72.57\\
Lunch room & $1\to 2$ & 47.53 & 62.09 & 48.50 & 62.13\\
Lunch room & $2\to 1$ & 46.31 & 60.98 & 49.64 & 62.93\\
Meeting room & $1\to 2$ & 56.78 & 71.68 & 64.29 & 77.52\\
Meeting room & $2\to 1$ & 52.98 & 68.33 & 63.62 & 76.19\\
Playground & $1\to 2$ & 52.41 & 67.13 & 57.25 & 70.78\\
Playground & $2\to 1$ & 50.52 & 65.21 & 56.37 & 69.29\\
Porch & $1\to 2$ & 74.08 & 84.64 & 82.10 & 89.54\\
Porch & $2\to 1$ & 63.13 & 76.29 & 69.84 & 80.60\\
Pots & $1\to 2$ & 73.12 & 84.38 & 87.35 & 93.19\\
Pots & $2\to 1$ & 55.72 & 71.16 & 74.60 & 85.13\\
Printing area & $1\to 2$ & 75.11 & 85.55 & 77.92 & 87.37\\
Printing area & $2\to 1$ & 59.79 & 74.29 & 64.07 & 77.33\\
Zen & $1\to 2$ & 67.62 & 76.30 & 75.37 & 81.07\\
Zen & $2\to 1$ & 59.47 & 71.76 & 69.67 & 78.91\\
\midrule
\multicolumn{2}{l}{Mean} & 59.93 & 73.39 & 67.15 & 78.41\\
\bottomrule
\end{tabular*}
\end{table}

\subsection{Fitted Priors and Change Area}
\label{app:prior-area}

We analyse the scene-specific priors fitted from reference edits.

\begin{figure}[t]
\centering
\includegraphics[width=0.88\linewidth]{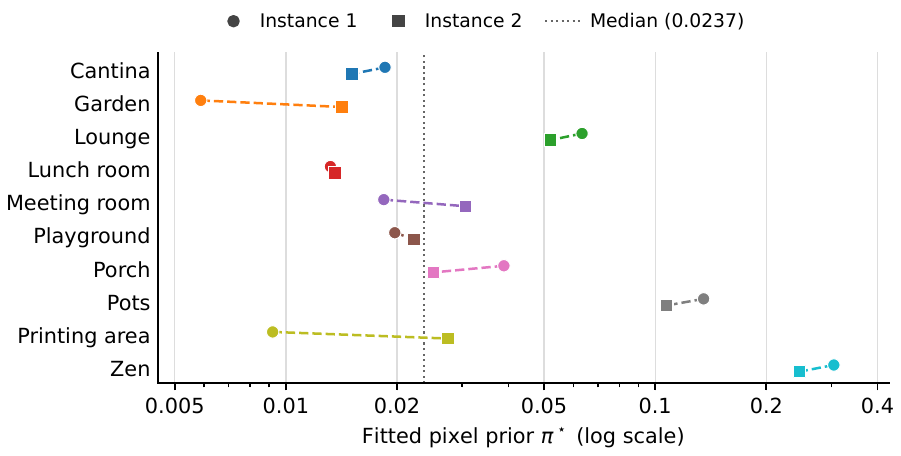}
\caption{Fitted pixel priors on PASLCD. Each scene has a distinct colour; circles and squares denote its two instances, joined by a dashed line. The vertical dotted line marks the median over all 20 instances. The horizontal axis uses a logarithmic scale.}
\label{fig:pixel-prior-distribution}
\end{figure}

Figure~\ref{fig:pixel-prior-distribution} shows the effective pixel priors $\pi^\star$ fitted from reference data for the 20 PASLCD instances.
They range from $0.00588$ to $0.30471$, with median $0.02368$ and interquartile range $[0.01487,0.05489]$.
Each prior selects an instance-specific decision operating point and remains fixed during online inference; it does not estimate the fraction of changed pixels in a query image.

To examine performance across change sizes, we group the 500 evaluated frames by the fraction of ground-truth changed pixels, keeping all fitted parameters and predictions unchanged (Table~\ref{tab:change-area}).
For each query, we average the metrics from the two complete runs, then average over queries within each group, with false-positive rate $\mathrm{FPR}=\mathrm{FP}/(\mathrm{FP}+\mathrm{TN})$ and false-negative rate $\mathrm{FNR}=\mathrm{FN}/(\mathrm{TP}+\mathrm{FN})$.
Frames with less than $1\%$ change have lower mIoU and higher FNR than the $1$--$5\%$ group; refinement improves their mIoU while slightly increasing FNR.
The observed change fractions span $0.17$--$20.13\%$, with only one frame above $20\%$ and none at $30\%$ or higher.
Scene and viewpoint composition differ across groups, so this analysis describes fixed-prior performance rather than isolating the effect of change area or establishing robustness to prior drift.

\begin{table}[t]
\centering
\small
\caption{PASLCD performance by ground-truth change area. Metrics average two runs over the same 500 queries (\%). The $20$--$30\%$ group contains only one frame; no evaluated frame has $\geq30\%$ change area.}
\label{tab:change-area}
\setlength{\tabcolsep}{4pt}
\begin{tabular*}{\linewidth}{@{\extracolsep{\fill}}lrr rrr rrr@{}}
\toprule
 & & & \multicolumn{3}{c}{Pixel output} & \multicolumn{3}{c}{Refined output}\\
\cmidrule(lr){4-6}\cmidrule(l){7-9}
Change area & Frames & Instances & mIoU$\uparrow$ & FPR$\downarrow$ & FNR$\downarrow$ & mIoU$\uparrow$ & FPR$\downarrow$ & FNR$\downarrow$\\
\midrule
$<1\%$ & 88 & 12 & 49.50 & 0.64 & 36.13 & 54.83 & 0.58 & 37.74\\
$1$--$5\%$ & 298 & 18 & 61.18 & 0.96 & 18.18 & 70.11 & 0.60 & 16.69\\
$5$--$10\%$ & 84 & 12 & 66.39 & 0.97 & 24.88 & 71.12 & 0.67 & 22.48\\
$10$--$20\%$ & 29 & 4 & 67.35 & 1.67 & 25.27 & 71.10 & 1.44 & 22.38\\
$20$--$30\%$ & 1 & 1 & 63.44 & 5.19 & 23.50 & 66.87 & 4.82 & 20.33\\
\bottomrule
\end{tabular*}
\end{table}

\subsection{Relighting Strength}
\label{app:relighting-strength}

We vary the strength of the reference-fitted colour perturbation while retaining the object edits, labels, camera subsets, and query order.
For each selected coefficient vector $\theta$, we use $\theta_K=\theta_{\mathrm{id}}+K(\theta-\theta_{\mathrm{id}})$, where $\theta_{\mathrm{id}}$ gives the identity colour transform, and render the transformed Gaussian colours with the same clipping rule.
The default is $K=1$; $K=0$ removes the relighting perturbation, and $K=2$ doubles it.
At $K=0$, the relighting sample slots are retained with identity coefficients, so all settings contain the same 11,384 training records.
For $K=0$ and $K=2$, we refit suppression, the likelihood-ratio mapping, unchanged-response statistics, image weights, the pixel prior, and the contour filter using only reference data and the same fitting settings.
Table~\ref{tab:relighting-strength} reports pixel and full outputs on the same 20 PASLCD instances and 500 query frames, with $K=1$ reusing the main results.

Default relighting improves full-output mIoU by $0.66$ points over $K=0$, with gains in 13 of 20 instances.
Doubling the strength changes pixel and full mIoU by $+0.03$ and $-0.07$ points, respectively, relative to the default.
These results indicate a modest benefit from the adopted augmentation, with similar aggregate performance at $K=1$ and $K=2$.

\begin{table}[t]
\centering
\caption{Relighting strength on PASLCD, averaged over 20 instances (\%). All settings retain the same training records; $K$ scales the lighting perturbation about the identity transform.}
\label{tab:relighting-strength}
\small
\begin{tabular}{lrrrr}
\toprule
& \multicolumn{2}{c}{Pixel output} & \multicolumn{2}{c}{Full output}\\
\cmidrule(lr){2-3}\cmidrule(lr){4-5}
Strength & mIoU $\uparrow$ & F1 $\uparrow$ & mIoU $\uparrow$ & F1 $\uparrow$\\
\midrule
$K=0$ & 60.04 & 73.54 & 66.98 & 78.45\\
$K=1$ (default) & 60.36 & 73.80 & \textbf{67.64} & \textbf{78.87}\\
$K=2$ & \textbf{60.40} & \textbf{73.83} & 67.57 & 78.85\\
\bottomrule
\end{tabular}
\end{table}

\subsection{Construction of Historical Support}
\label{app:history-rules}

We compare two ways of constructing historical support: our Bayes-factor accumulation and temporal averaging, using pixel outputs on the same 20 PASLCD instances and 500 frames.
Temporal averaging takes the mean of each Gaussian's valid responses $r_{tg}=[\mathcal A_t(C_t)]_g$ across earlier frames, selected by $\mathbb I_{tg}$, and assigns support $0.5$ to unobserved Gaussians.
These values are rendered to obtain $H_t$, with prediction preceding the current update as in our method.
We retain the image-evidence pipeline and query order, and refit the pixel prior on the editing subset using the same fitting procedure, without target labels.
Table~\ref{tab:history-rules} reuses the no-history and full-pixel results from Table~\ref{tab:component-ablation}.
Our construction improves mean mIoU by $4.42$ points over temporal averaging and performs better in 16 of 20 instances.

\begin{table}[t]
\centering
\caption{Historical support on PASLCD, averaged over 20 instances (\%, higher is better). Ours uses Bayes-factor accumulation. All results use pixel outputs without refinement.}
\label{tab:history-rules}
\small
\begin{tabular}{lrr}
\toprule
Historical support & mIoU $\uparrow$ & F1 $\uparrow$\\
\midrule
None & 57.81 & 71.67\\
Temporal averaging & 55.94 & 70.27\\
Ours & \textbf{60.36} & \textbf{73.80}\\
\bottomrule
\end{tabular}
\end{table}

\subsection{Evaluation and Runtime Details}
\label{app:updated-evaluation}

\textbf{Evaluation and repeatability.}
Queries are processed from raw images, with pose estimation and feature extraction performed online. The main PASLCD results (Tables~\ref{tab:overall} and~\ref{tab:per-scene}) and ablations (Tables~\ref{tab:component-ablation}, \ref{tab:relighting-strength}, and~\ref{tab:history-rules}) average two complete runs over all 500 queries per configuration. For the full method, both runs achieve $67.64\%$ mIoU; the largest per-instance difference is $0.019$ percentage points. CCS3D uses all 18 queries across four scenes at $2\times$ downsampling. Binary predictions are resized to the union masks by nearest-neighbour interpolation; Table~\ref{tab:ccs3d} averages per-scene metrics equally over two runs.

\textbf{Continuous sequence timing.}
After initialization, synchronized wall-clock timing starts before the first query is retrieved and ends when the final mask is ready. Input prefetching starts during initialization for both methods; remaining input waits and all inter-frame work are included. O-SCD also waits for GPU work to complete at its stage synchronization points; these waits are included in the timed sequence. Mask export and metric evaluation occur outside timing. FPS is the total number of processed frames divided by summed sequence duration: 1,000 frames across two runs for each of our output modes and 500 frames for O-SCD(O). The pixel runs achieve $27.88$ and $27.98$\,FPS; the full runs achieve $17.15$ and $17.19$\,FPS.

\textbf{Batch refinement timing.}
O-SCD(R) first processes the query images sequentially, then jointly refines its change representation using all views and renders the final masks. Its timing covers this complete procedure, from the first input retrieval to the last refined mask, excluding initialization and file export. We divide total sequence time by the number of frames for Table~\ref{tab:runtime}; the resulting per-frame time and equivalent FPS describe batch processing, not online prediction latency. Over the same 20 instances, complete processing averages $5.03$\,s per 25-frame sequence, including $3.29$\,s of post-sequence optimization and $6.48$\,ms of final mask rendering.

\textbf{Initialization.}
Initialization comprises loading scene and model parameters, preparing reference matching data, and pre-executing model and GPU operations using non-test inputs. Both methods run SAM on all-zero inputs. Our method additionally processes a fixed reference-edit observation twice through the complete detection pipeline, using synthetic data already available from offline editing. We then restore the detector history, reference matching state, and random-generator states before processing the query sequence. Initialization averages $14.08$\,s for our pixel pipeline, $15.54$\,s for our full pipeline, and $7.16$\,s for O-SCD(O). The two reference-edit passes take $0.111$ and $0.168$\,s for the pixel and full pipelines. Mean first-frame latency is $35.55$, $58.24$, and $110.38$\,ms, respectively.

\textbf{Stage-wise timing.}
The frontend comprises query relocalisation, reference rendering, and initial change-cue computation (Section~\ref{sec:preliminaries}). Post-frontend time covers detection, temporal-state updates, and refinement where applicable. Table~\ref{tab:runtime-stages} averages synchronized profiles over all 25 frames of all 20 instances. Detection includes the pixel decision and history update; refinement includes its RGB transfer. Input waits and inter-frame work are reported separately. Our profiles use a separate pass from the continuous throughput measurements in Table~\ref{tab:runtime}; O-SCD(O) uses its synchronized sequence records.

\begin{table}[t]
\centering
\caption{Online runtime breakdown on PASLCD (RTX 5090; 20 instances, frames 1--25; ms/frame). Total sums all listed intervals. Post-frontend in Table~\ref{tab:runtime} sums detection and refinement.}
\label{tab:runtime-stages}
\small
\setlength{\tabcolsep}{5pt}
\begin{tabular*}{\linewidth}{@{\extracolsep{\fill}}lrrrrr@{}}
\toprule
Method & Frontend & Detection & Refinement & Input/loop & Total\\
\midrule
O-SCD(O) & 31.62 & 37.65 & 0.00 & 0.06 & 69.33\\
Ours (pixel) & 31.24 & 4.79 & 0.00 & 0.07 & 36.10\\
Ours (full) & 31.51 & 5.34 & 20.29 & 0.10 & 57.24\\
\bottomrule
\end{tabular*}
\end{table}

\subsection{Offline Preparation and Component Timing Breakdown}
\label{app:offline-timing}

Offline preparation includes fitting the pixel prior and contour filter from reference edits.

Table~\ref{tab:offline-timing} details the offline preparation and component fitting time across all 20 PASLCD scene-instances, measured on an NVIDIA RTX 5090 using wall-clock timing with GPU synchronization.
The offline workflow comprises two main phases: (1)~\emph{synthetic edit generation}, which extracts multi-view object proposals with SAM2, performs 3D Gaussian edits (recolouring, removal, translation), renders lighting-varied views, and forms training observation pairs; and (2)~\emph{scene component fitting}, which executes reference preprocessing, fits the pixel pipeline components (response suppression, log-likelihood ratio mapping, unchanged response moments, and 1D branch-and-bound prior search $\pi^\star$), and trains the 18-dimensional symmetric contour filter with Platt scaling.
Across all instances, the complete offline workflow averages $215.82$\,s ($\sim$3.6\,minutes; median $197.13$\,s, ranging from $142.15$\,s to $440.69$\,s), given an existing reference reconstruction.

\begin{table}[t]
\centering
\caption{Complete offline preparation and fitting runtime breakdown across 20 PASLCD scene-instances (seconds, measured on an NVIDIA RTX 5090). Generation includes automatic object proposals, 3DGS edits, and synthetic paired observation rendering. Generated records are counted before the camera split. Fitting covers reference preprocessing, pixel models, prior search, and contour filter fitting.}
\label{tab:offline-timing}
\small
\setlength{\tabcolsep}{4.5pt}
\begin{tabular}{llrrrrrrr}
\toprule
Scene & Inst. & Retained & Generated & Generation & Preprocess & Pixel & Contour & Total\\
& & objects & records & (s) $\downarrow$ & (s) $\downarrow$ & (s) $\downarrow$ & (s) $\downarrow$ & (s) $\downarrow$\\
\midrule
Cantina & 1 & 22 & 366 & 106.95 & 8.21 & 41.68 & 18.84 & 175.68\\
& 2 & 28 & 602 & 112.19 & 7.99 & 56.92 & 22.93 & 200.02\\
Garden & 1 & 37 & 1080 & 119.42 & 7.57 & 83.23 & 30.48 & 240.69\\
& 2 & 35 & 752 & 121.05 & 8.25 & 69.61 & 33.54 & 232.46\\
Lounge & 1 & 18 & 384 & 112.07 & 8.26 & 42.39 & 19.14 & 181.86\\
& 2 & 18 & 340 & 79.13 & 6.92 & 37.24 & 18.85 & 142.15\\
Lunch\_room & 1 & 56 & 1744 & 226.44 & 11.28 & 119.27 & 47.17 & 404.16\\
& 2 & 64 & 2034 & 231.67 & 11.16 & 131.89 & 65.96 & 440.69\\
Meeting\_room & 1 & 17 & 410 & 90.87 & 6.78 & 44.82 & 28.05 & 170.52\\
& 2 & 21 & 484 & 85.33 & 7.09 & 48.84 & 35.81 & 177.07\\
Playground & 1 & 31 & 718 & 114.45 & 7.85 & 66.88 & 50.15 & 239.34\\
& 2 & 33 & 642 & 117.92 & 7.68 & 63.83 & 74.38 & 263.81\\
Porch & 1 & 44 & 842 & 102.77 & 7.11 & 67.46 & 34.04 & 211.38\\
& 2 & 43 & 788 & 122.44 & 7.81 & 67.89 & 34.49 & 232.63\\
Pots & 1 & 14 & 280 & 74.38 & 7.95 & 38.18 & 22.04 & 142.55\\
& 2 & 20 & 400 & 75.64 & 7.44 & 46.23 & 23.70 & 153.02\\
Printing\_area & 1 & 23 & 756 & 107.85 & 7.50 & 62.35 & 22.07 & 199.77\\
& 2 & 23 & 416 & 85.96 & 6.76 & 41.17 & 19.43 & 153.33\\
Zen & 1 & 24 & 564 & 91.22 & 8.27 & 58.68 & 36.32 & 194.49\\
& 2 & 18 & 378 & 77.34 & 7.58 & 48.64 & 27.11 & 160.68\\
\midrule
Mean & --- & 29.45 & 699.0 & 112.75 & 7.97 & 61.86 & 33.23 & \textbf{215.82}\\
Median & --- & 23.50 & 583.0 & 107.40 & 7.74 & 57.80 & 29.26 & \textbf{197.13}\\
\bottomrule
\end{tabular}
\end{table}

\clearpage
\section{Additional Visual Results}
\label{app:visual-results}

\begin{figure}[htbp]
\centering
\includegraphics[width=\linewidth]{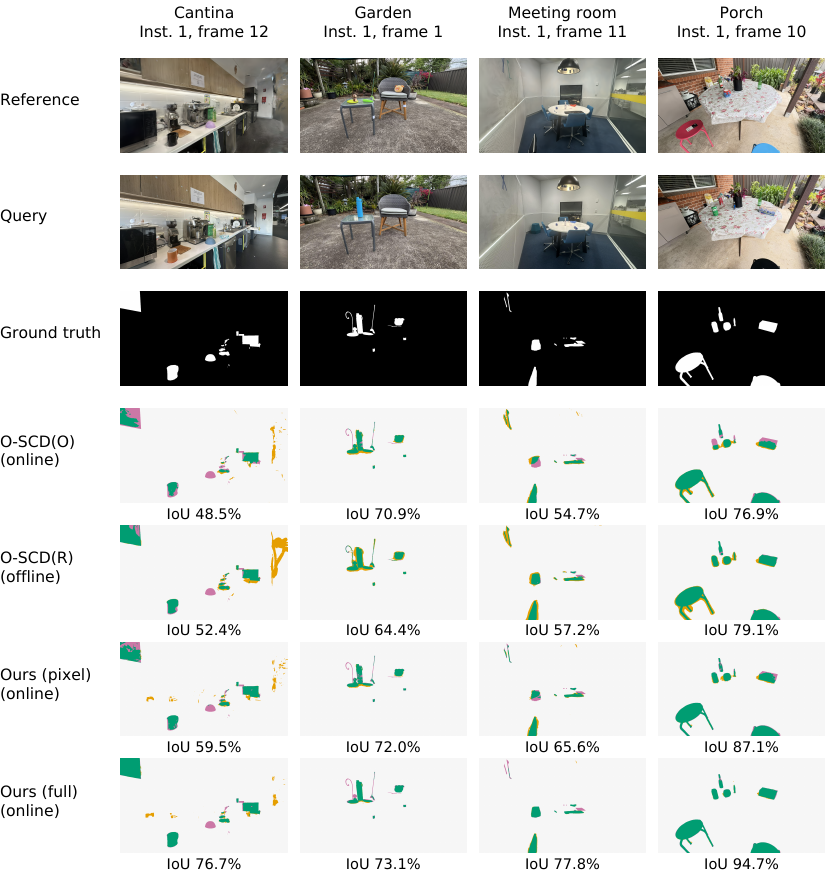}
\caption{\textbf{Qualitative comparison on PASLCD.}
Examples are from Cantina, Garden, Meeting room, and Porch.
Each column shows one query, with reference, query, ground truth, and method outputs arranged vertically.
Both O-SCD variants and our pixel and full outputs are shown.
Green, orange, and pink denote correct detections, false positives, and missed changes, respectively.
Numbers report per-frame foreground IoU. Thin structures remain partly missed in Garden.}
\label{fig:qualitative-additional}
\end{figure}

\clearpage
\begin{figure}[htbp]
\centering
\includegraphics[width=\linewidth]{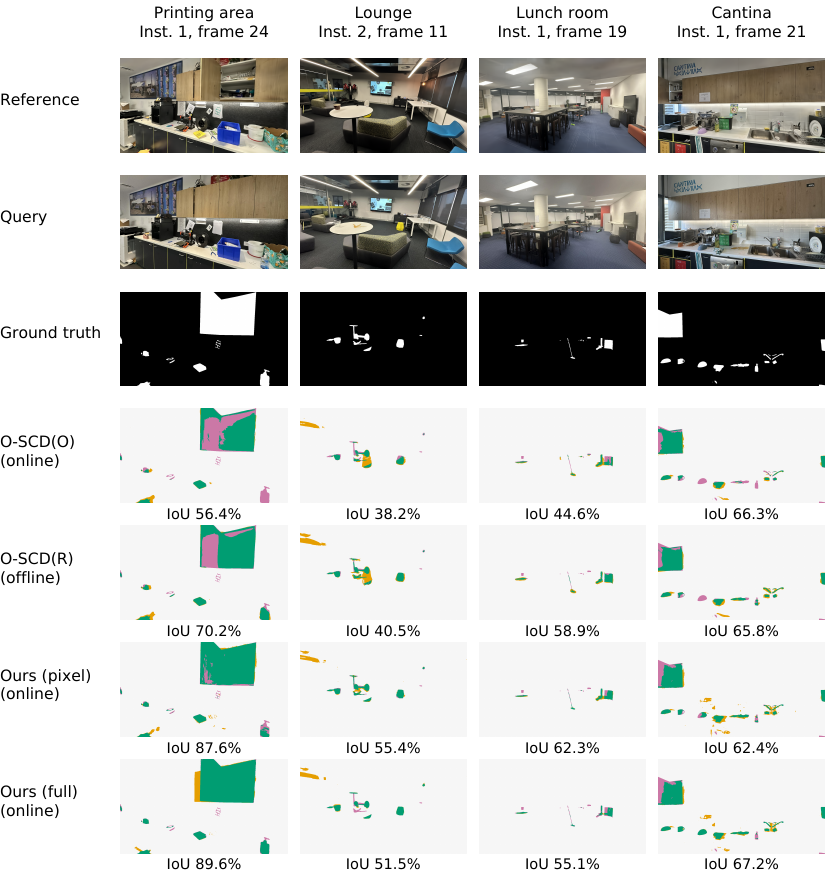}
\caption{\textbf{Additional qualitative comparisons on PASLCD.}
Examples are from Printing area, Lounge, Lunch room, and another view of Cantina.
The layout and colours follow Figure~\ref{fig:qualitative-additional}.
Refinement recovers the cabinet region in Printing area, while false positives near the ceiling in Lounge and missed object regions in Lunch room remain.
Numbers report per-frame foreground IoU.}
\label{fig:qualitative-additional-2}
\end{figure}

\clearpage
\subsection{Comparison on CCS3D}

\begin{figure}[htbp]
\centering
\includegraphics[width=\linewidth]{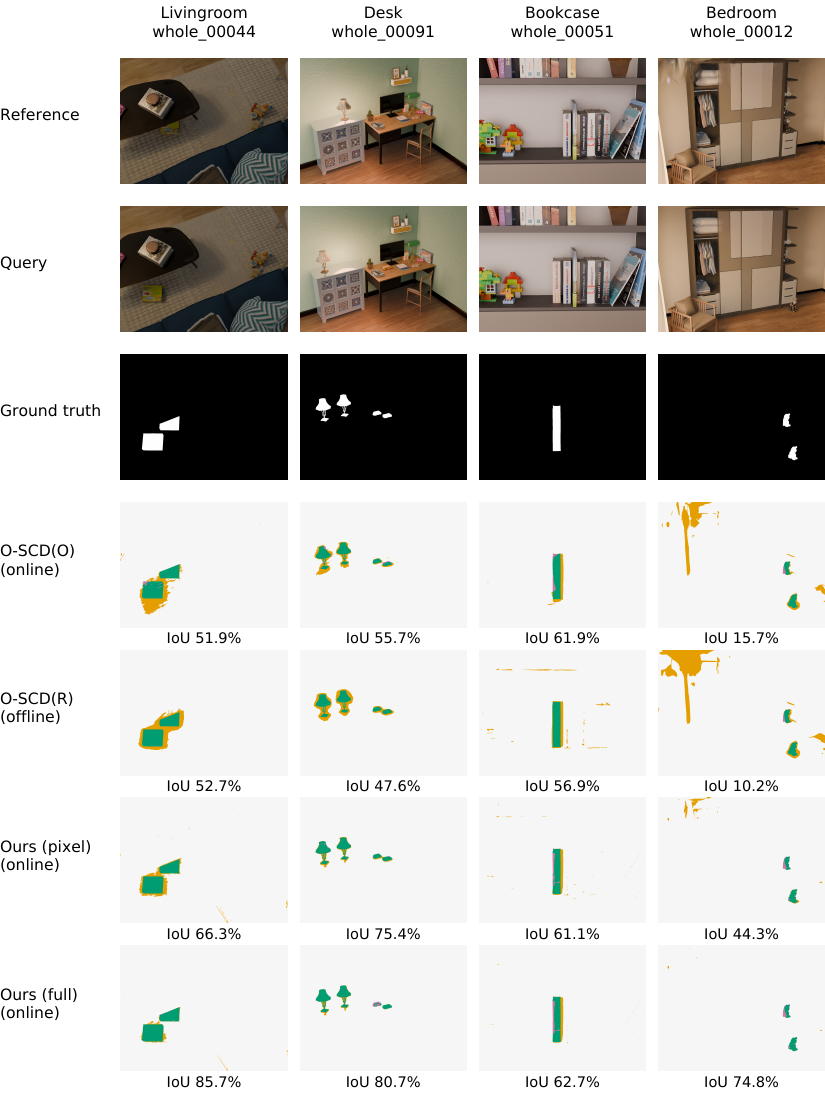}
\caption{\textbf{Qualitative comparison on CCS3D.}
Examples are from Livingroom, Desk, Bookcase, and Bedroom.
Rows and colours follow Figure~\ref{fig:qualitative-additional}.
Ground truth is the union of pre- and post-state masks, as in Table~\ref{tab:ccs3d}.
Our full method reduces excess responses around displaced objects in Livingroom and Desk and on unchanged regions in Bedroom; boundary errors remain in Bookcase.
Numbers report per-frame foreground IoU.}
\label{fig:qualitative-ccs3d}
\end{figure}

\clearpage
\subsection{Evidence Fusion and Component Ablations}

\begin{figure}[!htbp]
\centering
\includegraphics[width=\linewidth]{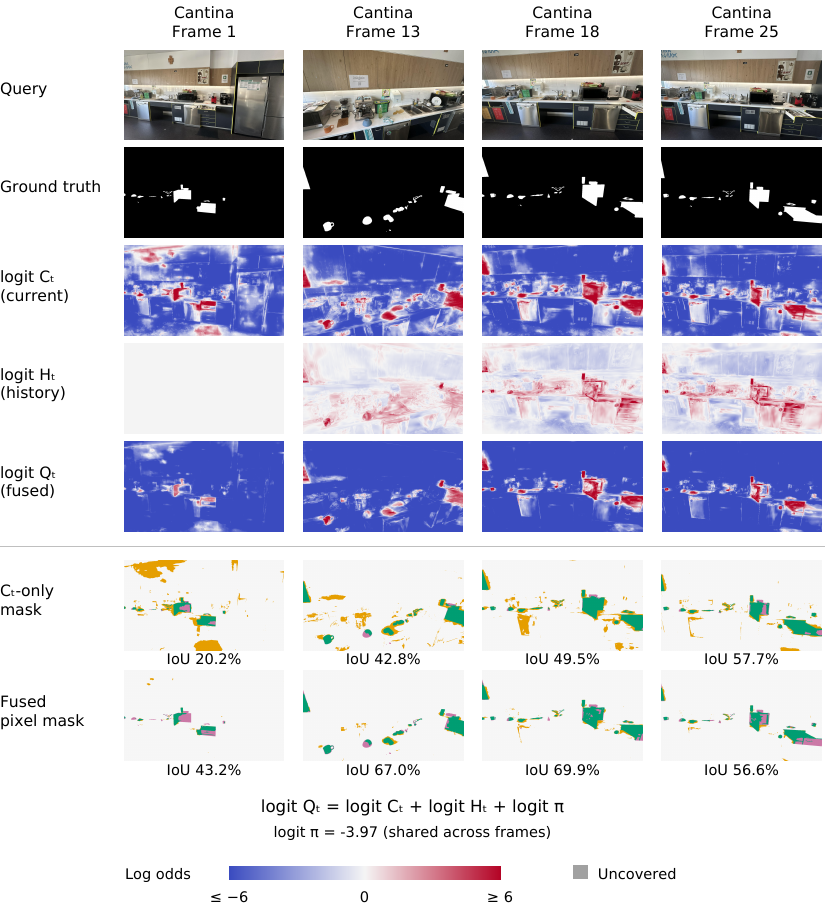}
\caption{\textbf{Evidence fusion in Cantina.}
Columns show frames 1, 13, 18, and 25 from one sequence.
Log-odds maps share a colour scale saturated at $\pm6$ for display; white denotes zero and grey marks uncovered pixels.
The $C_t$-only mask sets $H_t=\pi=0.5$, giving $Q_t^{\mathrm{pixel}}=C_t$.
Both masks threshold $Q_t^{\mathrm{pixel}}$ at $0.5$; fusion combines the same $C_t$, historical support $H_t$ from $F_{t-1}$, and the fitted prior $\pi$ (Eq.~\ref{eq:final-support}).
Mask colours follow Figure~\ref{fig:qualitative-additional}; numbers report per-frame foreground IoU.}
\label{fig:qualitative-history}
\end{figure}

\begin{figure}[htbp]
\centering
\includegraphics[width=\linewidth,height=0.86\textheight,keepaspectratio]{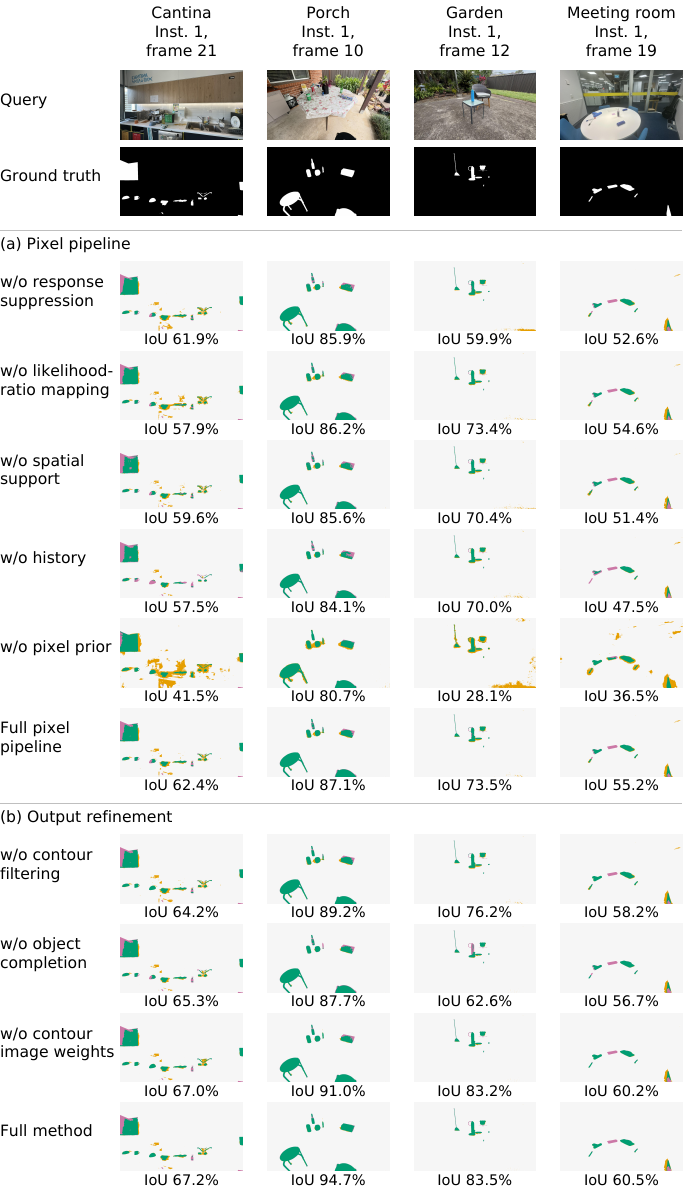}
\caption{\textbf{Qualitative component ablations.}
Examples are from Cantina, Porch, Garden, and Meeting room in PASLCD.
(a) Five pixel-pipeline ablations and the full pixel pipeline. The no-prior variant uses $\pi=0.5$; spatial support denotes Gaussian-aggregated support $S_t$.
(b) Three output-refinement ablations and the full method. The image-weight ablation omits image weighting during contour fitting.
Both groups use the same queries. Colours follow Figure~\ref{fig:qualitative-additional}; numbers report per-frame foreground IoU.}
\label{fig:qualitative-ablations}
\end{figure}

Figure~\ref{fig:qualitative-history} illustrates evidence fusion. Current-view evidence $C_t$ gives strong responses around candidate changes, including some unchanged regions. Historical support $H_t$ adjusts these responses using earlier observations, while the scene-fitted prior $\pi$ determines the evidence threshold for predicting change. The fused pixel probability $Q_t^{\mathrm{pixel}}$ reduces false positives but can also suppress true changes, as frame 25 illustrates. Across this 25-frame sequence, mean IoU increases from $47.31\%$ with $C_t$ alone to $57.84\%$ after fusion.
Figure~\ref{fig:qualitative-ablations} complements the quantitative component ablations in Table~\ref{tab:component-ablation}.

\clearpage

\end{document}